\documentclass{article} 
\usepackage{iclr2025_conference,times}

\usepackage{microtype}
\usepackage{hyperref}
\usepackage{url}
\usepackage{booktabs}

\usepackage{bm}
\usepackage{graphicx}

\usepackage{amsmath,amssymb,amsfonts}
\usepackage{courier}
\usepackage{color}
\usepackage{amsthm}
\usepackage{hyperref}
\usepackage{subfigure}
\usepackage{enumitem}
\usepackage{bbm}
\usepackage{wrapfig}
\usepackage{makecell}
\usepackage{multirow}
\usepackage[ruled,vlined]{algorithm2e}

\usepackage{booktabs}
\usepackage{pifont}
\usepackage{colortbl}
\usepackage[font=small]{caption}
\definecolor{mygraylite}{gray}{.94}
\definecolor{mygray}{gray}{.89}
\definecolor{darkergreen}{RGB}{21, 152, 56}
\definecolor{amber}{rgb}{1.0, 0.75, 0.0}
\definecolor{darkseagreen}{rgb}{0.56, 0.74, 0.56}
\definecolor{darkblue}{rgb}{0, 0, 54.5}
\definecolor{darkorange}{rgb}{220,88,42}
\newcommand*{\modelname}{\text{LLM-SR}\@\xspace}

\newcommand{\changed}[2]{\textcolor{black}{#1}}

\newcommand{\authorsep}{\hspace{2ex}}
\newcommand{\instsep}{\hspace{2ex}}

\newlist{myitemize2}{itemize}{4}
\setlist[myitemize2,1]{label={},leftmargin=0em}
\newlist{myitemize}{itemize}{4}
\setlist[myitemize,1]{label=\textbullet,leftmargin=1em}
\usepackage{hyperref}
\usepackage{url}

\usepackage{listings}
\usepackage{xcolor}

\definecolor{codegreen}{rgb}{0,0.6,0}
\definecolor{codegray}{rgb}{0.5,0.5,0.5}
\definecolor{codepurple}{rgb}{0.58,0,0.82}
\definecolor{backcolour}{rgb}{0.95,0.95,0.92}
\definecolor{inlinebackcolour}{rgb}{0.95,0.95,0.95}

\lstdefinestyle{customstyle}{
    basicstyle=\linespread{1.0}\ttfamily\nornmalsize,
    breakatwhitespace=false,
    breaklines=true,
    captionpos=b,
    keepspaces=true,
    numbers=left,
    numbersep=5pt,
    showspaces=false,
    showstringspaces=false,
    showtabs=false,
    tabsize=4,
    frame=single,
    framesep=2mm,
    backgroundcolor=\color{backcolour},
    commentstyle=\color{codegreen},
    keywordstyle=\color{blue},
    stringstyle=\color{codepurple},
}

\lstdefinestyle{inlinestyle}{
    basicstyle=\ttfamily\small,
    breakatwhitespace=false,
    breaklines=true,
    keepspaces=true,
    language=Python,
    backgroundcolor=\color{inlinebackcolour},
    frame=none,
    framerule=0pt,
    framesep=1pt,
    framexleftmargin=3pt,
    framexrightmargin=3pt,
    framextopmargin=1pt,
    framexbottommargin=1pt,
    keywordstyle=\color{green!60!black}\bfseries,
    identifierstyle=\color{blue},
}

\usepackage{adjustbox} 
\usepackage{multicol}

\title{\modelname: Scientific Equation Discovery via \\Programming with Large Language Models}

\author{Parshin Shojaee$^{1\thanks{Equal contribution. Contact: parshinshojaee@vt.edu, mmeidani@andrew.cmu.edu}}$ \authorsep
Kazem Meidani$^{2^{*}}$  \authorsep Shashank Gupta$^{3}$ \\
\textbf{Amir Barati Farimani}$^{2}$
\authorsep 
\textbf{Chandan K. Reddy}$^1$
\\\\
$^{1}$Virginia Tech \instsep
$^{2}$Carnegie Mellon University \instsep
$^{3}$Allen Institute for AI
}

\iclrfinalcopy

\begin{document}

\maketitle

\begin{abstract}
Mathematical equations have been unreasonably effective in describing complex natural phenomena across various scientific disciplines. However, discovering such insightful equations from data presents significant challenges due to the necessity of navigating extremely large combinatorial hypothesis spaces. Current methods of equation discovery, commonly known as symbolic regression techniques, largely focus on extracting equations from data alone, often neglecting the domain-specific prior knowledge that scientists typically depend on. 
They also employ limited representations such as expression trees, constraining the search space and expressiveness of equations. To bridge this gap, we introduce LLM-SR, a novel approach that leverages the extensive scientific knowledge and robust code generation capabilities of Large Language Models (LLMs) to discover scientific equations from data. Specifically, LLM-SR treats equations as programs with mathematical operators and combines LLMs' scientific priors with evolutionary search over equation programs. The LLM iteratively proposes new equation skeleton hypotheses, drawing from its domain knowledge, which are then optimized against data to estimate parameters. 
We evaluate LLM-SR on four benchmark problems across diverse scientific domains (e.g., physics, biology), which we carefully designed to simulate the discovery process and prevent LLM recitation. 
Our results demonstrate that LLM-SR discovers physically accurate equations that significantly outperform state-of-the-art symbolic regression baselines, particularly in out-of-domain test settings. \changed{We also show that LLM-SR's incorporation of scientific priors enables}{LLM-SR's incorporation of scientific priors also enables} more efficient equation space exploration than the baselines\footnote[1]{Code and data are available: \url{https://github.com/deep-symbolic-mathematics/LLM-SR}}.
\end{abstract}

\section{Introduction}
\label{sec:intro}
\vspace{-0.5em}
The emergence of Large Language Models (LLMs) has marked a significant milestone in artificial intelligence, showcasing remarkable capabilities across various domains \citep{achiam2023gpt}. As LLMs continue to evolve, researchers are exploring innovative ways to harness their potential for solving complex problems such as scientific discovery \citep{scientific-age-AI,llm-scientific-gpt-microsoft}.
Their ability to process and comprehend vast amounts of scientific literature, extract relevant information, and generate coherent hypotheses has recently opened up new avenues for accelerating scientific progress \citep{LLM-scientific-synthesis, ji2024scimon}.
Additionally, by leveraging their ability to understand and reason with the help of programming and execution, LLMs have shown the potential to enhance automated reasoning and problem-solving capabilities for general natural language and mathematics optimization tasks, e.g., prompt optimization and heuristic discovery~\citep{LMX-Lehman-2023,LLM-Optimizer,self_refine-2024,FunSearch}.
Motivated by these strengths, LLMs could be particularly helpful for the task of equation discovery, a fundamental task in science and scientific discovery.

Discovering accurate symbolic mathematical models from data is an important task in various scientific and engineering disciplines. 
The task of \emph{data-driven equation discovery} (also commonly known as Symbolic Regression (SR)), aims to find abstract mathematical equations 
from data observations such that these equations are predictive of the underlying data, are interpretable, and generalize to unseen data from the same physical phenomena. 
Finding such equations offers several advantages over simply estimating a predictive model, as the resulting mathematical functions provide insights into the underlying physical processes, enable extrapolation beyond the observed data, and facilitate knowledge transfer across related problems \citep{langley1981data, Schmidt-Lipson-2009}.
However, while evaluating the fit of a proposed equation is relatively straightforward, the inverse process of obtaining these mathematical equations from data is a challenging problem, known to be NP-hard \citep{NP-hard-Symbolic-2022}. 
Current equation discovery methods encompass a wide variety of approaches from evolutionary search algorithms \citep{pysr-2023, mundhenk-seeding-GP-NeurIPS-2021, SRBench-Cava-NeurIPS-2021} to advanced deep learning methods using Transformers
\citep{Biggio-NeSymReS-ICML-2021,Kamienny-E2E-symbolic-NIPS-2022}. 
Most of the traditional symbolic regression techniques are built on top of Genetic Programming~(GP) \citep{Koza1994-GP} evolutionary methods, representing mathematical equations as expression trees and searching the combinatorial space of possible equations through iterative mutation and recombination.
However, these methods often struggle with the complexity of the vast optimization space and do not incorporate prior scientific knowledge, which leads to suboptimal solutions and inefficient exploration of the equation search space. 
Similarly, the design of current general LLM-based optimization frameworks \citep{LMX-Lehman-2023,FunSearch,LLM-Optimizer} also have several key limitations in terms of domain knowledge integration and diverse exploration which are critical for equation discovery.
Thus, \textit{there is a need for specialized equation discovery methods that effectively integrate prior scientific knowledge into the navigation of vast equation search space, a strategy akin to a scientist’s reliance on foundational scientific knowledge when formulating hypotheses for scientific discovery.}

\begin{figure}[t]
\centering
\includegraphics[width=\linewidth]{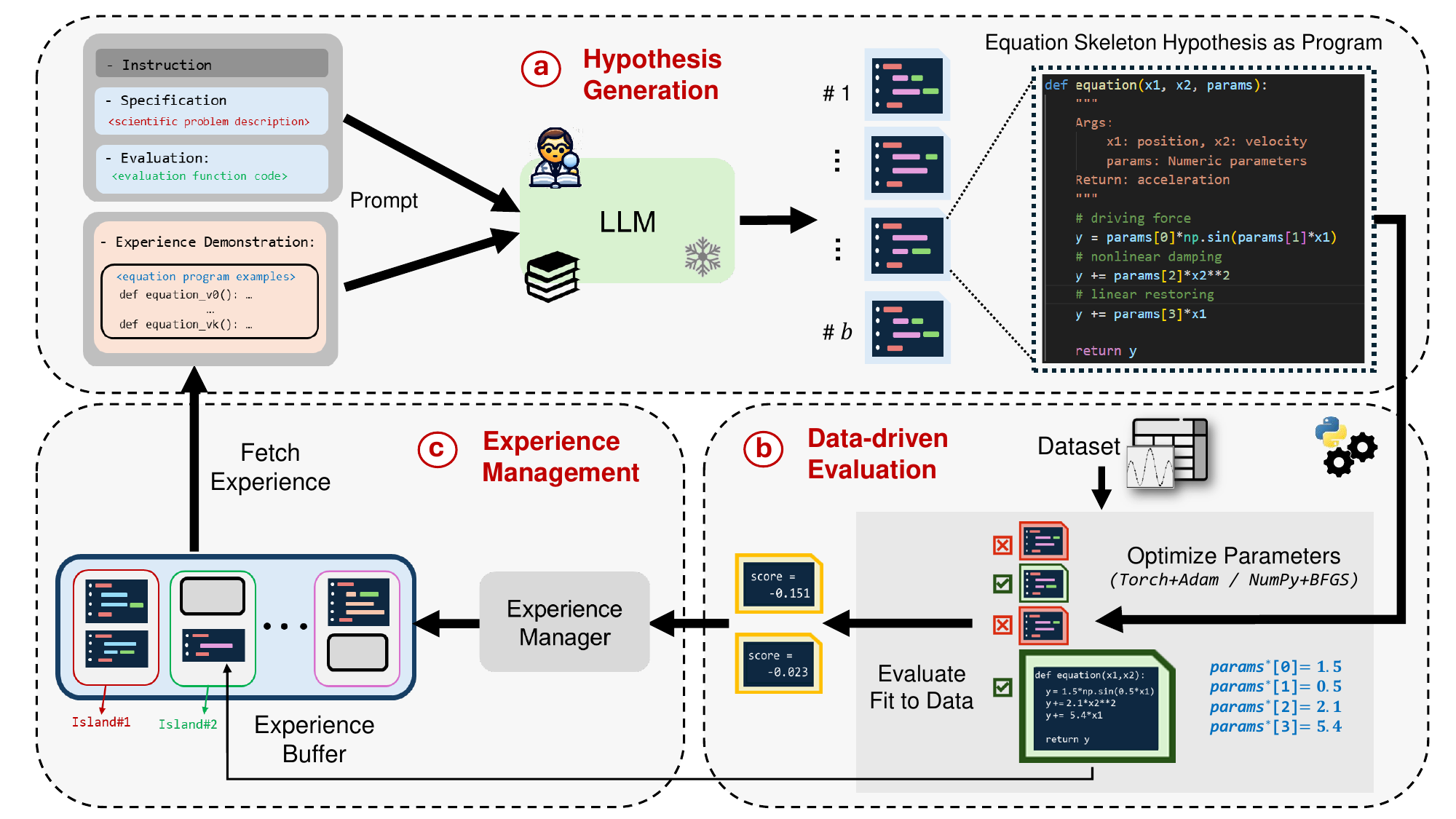}
\caption{
\textbf{The \modelname framework}, consisting of three main steps: \textbf{(a)} \textbf{Hypothesis Generation}, where LLM generates equation program skeletons based on a structured prompt;
\textbf{(b)} \textbf{Data-driven Evaluation}, which optimizes the parameters of each equation skeleton hypothesis and assesses its fit to the data; and \textbf{(c)} \textbf{Experience Management}, which maintains a diverse buffer of high-scoring hypotheses to provide informative in-context examples into LLM's prompt for effective iterative refinement.
\vspace{-0.5em}
}
\label{fig:llm-sr}
\vspace{-0.5em}
\end{figure}

To address all these limitations, we introduce \modelname (shown in Fig.~\ref{fig:llm-sr}), a novel framework that combines the strengths of LLMs, reliable optimizers, and evolutionary search for data-driven equation discovery.
At its core, \modelname is an iterative hypotheses refinement method that generates, evaluates, and refines equation hypotheses based on data-driven feedback.
Specifically, \modelname first prompts the LLM to propose new equation hypotheses~(Fig.~ \ref{fig:llm-sr}(a)), then evaluates their fit on the observed data using off-the-shelf optimizers~(Fig.~ \ref{fig:llm-sr}(b)), and uses this data-driven feedback and a carefully maintained dynamic memory of previous equations~(Fig.~ \ref{fig:llm-sr}(c)) to iteratively guide the search towards better equations.
\modelname leverages the scientific knowledge embedded in LLMs using short descriptions of the problem and the variables involved in a given system to generate educated hypotheses for equation skeletons (i.e., mathematical structures with placeholder parameters for numeric coefficients and constants). The LLM's in-context learning and crossover capabilities \citep{LMX-Lehman-2023}
are then employed to refine the suggested equation skeletons in an iterative process. 
By representing equations as Python programs, we take advantage of LLM's ability to generate structured and executable code \citep{starcoder, rl_code1} while providing a flexible and effective way to represent general mathematical relations. 
The program representation also facilitates direct and differentiable parameter optimization to better optimize the coefficients or constants in the generated equations. 

To leverage LLM's scientific prior knowledge yet prevent the risk of LLM recitation \citep{wu2023reasoning} in equation discovery (observed for common benchmarks like Feynman \citep{AI-Feynman-Science-2020}), we designed four custom benchmark problems across physics, biology, and materials science for the evaluation of \modelname. By incorporating synthetic modifications to physical models and experimental datasets, these problems aim to simulate the real discovery processes (see App.~\ref{sec:app-feyn} and \ref{sec:app-data} for details).
We evaluated \modelname using GPT-3.5-turbo \citep{brown2020language} and Mixtral-8x7B \citep{mixtral-llm} as backbone LLMs.
Results demonstrate that \modelname consistently outperforms state-of-the-art symbolic regression methods, discovering physically accurate equations with better fit and generalization in both in-domain (ID) and out-of-domain (OOD) test settings. 
By leveraging the scientific prior knowledge, \modelname explores the equation search space more efficiently, requiring fewer iterations to find accurate equations.
Our ablation analysis also highlights the crucial role of data-driven feedback, iterative refinement, and program representation in \modelname's performance.
The major contributions of this work are as follows:

\vspace{-0.6em}
\begin{itemize}[leftmargin=*]
\vspace{-0.1em}
\item We introduce \modelname, a novel framework that leverages domain-specific prior knowledge and code generation capabilities of LLMs combined with off-the-shelf optimizers and evolutionary search for data-driven scientific equation discovery.

\vspace{-0.1em}
\item We create four benchmark problems spanning physics, biology, and materials science, designed to simulate real-world discovery and prevent LLM recitation risks for evaluation of \modelname.

\vspace{-0.1em}
\item We show that \modelname outperforms state-of-the-art symbolic regression methods by navigating the equation search space more efficiently and discovering more accurate equations with better out-of-domain generalization.

\vspace{-0.1em}
\item We demonstrate through a comprehensive ablation study that natural language problem descriptions, program representation, data-driven feedback, and iterative hypothesis refinement are all essential components for \modelname's success.




\vspace{-0.1em}
\end{itemize}
\vspace{-0.5em}

\vspace{-0.5em}
\section{LLM-SR Methodology}
\vspace{-0.8em}
\subsection{Problem Formulation}
\label{sec:method-form}
\vspace{-0.8em}
In the task of data-driven equation discovery, also known as symbolic regression~(SR), the goal is to find a concise symbolic expression $\tilde{f}$ approximating an unknown function $f: \mathbb{R}^d \rightarrow \mathbb{R}$. Given a dataset $\mathcal{D} = \{(\mathbf{x}_i, y_i)\}_{i=1}^n$, SR methods seek to uncover the hidden mathematical relationship such that $\tilde{f}(\mathbf{x}_i) \approx y_i$, $\forall i$. The discovered equation should not only accurately fit the observed data points but also exhibit strong generalization capabilities to unseen data while maintaining interpretability.

Current SR methods typically represent equations using techniques such as expression trees \citep{pysr-2023}, prefix sequences \citep{DSR-Petersen-ICLR-2021,Biggio-NeSymReS-ICML-2021}, or context-free grammars \citep{Todorovski-Prob-grammar-SR}.
These representations provide structured and constrained search spaces, enabling evolutionary algorithms like genetic programming to explore and find candidate expressions. In contrast, we employ program functions to directly map inputs $\bm{x}$ to targets $\bm{y}$:  
\lstinline[language=Python, style=inlinestyle]{def f(x): ... return y}. This approach offers greater expressiveness in mathematical relations but expands the search space significantly.
To navigate this vast program space effectively, we leverage LLMs for their scientific knowledge and code generation capabilities. Let $\pi_\theta$ denote a pre-trained LLM with parameters $\theta$. We iteratively sample equation program skeletons $\mathcal{F}=\{f:f\sim\pi_\theta\}$, aiming to maximize the reward $\text{Score}_\mathcal{T}(f, \mathcal{D})$ for a given scientific problem $\mathcal{T}$ and dataset $\mathcal{D}$: $f^* = \arg\max_{f} \mathbb{E}_{d\in\mathcal{D}}\left[\text{Score}_\mathcal{T}(f, \mathcal{D})\right]$.
Our approach, \modelname, prompts the LLM to propose hypotheses based on problem specifications and demonstrations of previously discovered promising equations. The LLM generates equation program skeletons with placeholder parameters, which are then optimized using robust Python optimizers. Promising hypotheses are added to a dynamic experience buffer, guiding subsequent in-context example updates and equation refinement. Below we explain the key components of this framework, shown in Fig.~\ref{fig:llm-sr}, in more detail.

\vspace{-0.5em}
\subsection{Hypothesis Generation}
\label{sec:method-hyp}
\begin{figure}[t]
\centering
\includegraphics[width=0.85\linewidth]
{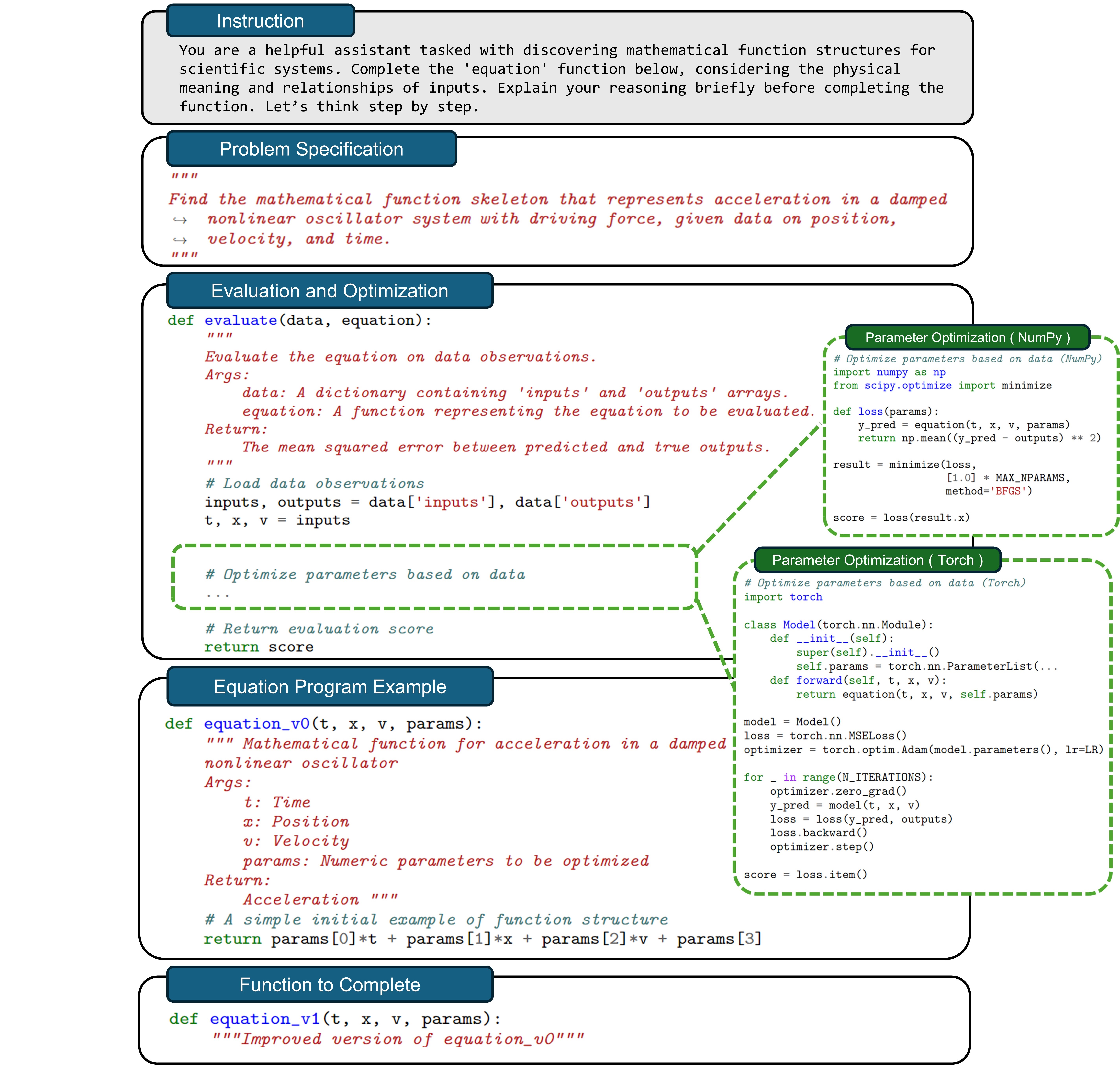}
\caption{\small Example of initial input prompt for the nonlinear oscillator discovery task, including problem specification, evaluation and optimization function, and the initial input equation example.
\vspace{-0.5em}
}
\vspace{-0.5em}
\label{fig:spec-bacteria}
\end{figure}

\vspace{-0.8em}
The hypothesis generation step (Fig.~\ref{fig:llm-sr}(a)) utilizes a pre-trained LLM to propose diverse and promising equation program skeletons. Our prompt structure, shown in Fig.~\ref{fig:spec-bacteria}, consists of the following components: \textbf{Instruction}: A clear directive for completing the function body, emphasizing consideration of physical meanings and relationships among input variables; \textbf{Problem Specification}: A concise description of the scientific problem, including key variables, constraints, and objectives; \textbf{Evaluation and Optimization Function}: The function used to assess the data-driven quality and fitness of proposed equations after parameter optimization; and \textbf{Experience Demonstration}: In-context examples of equation skeleton programs and their improvement trajectory.

At each iteration $t$, we sample a batch of $b$ equation skeletons $\mathcal{F}_t = \{f_i\}_{i=1}^b$ from the LLM $\pi_\theta$: $f_i \sim \pi_\theta(\cdot| \mathbf{p}_t)$ where $\mathbf{p}_t$ is the constructed prompt. We employ stochastic temperature-based sampling to balance exploration (creativity) and exploitation (prior knowledge) in the hypothesis space. Sampled equation programs are executed, and those failing to execute or exceeding a maximum execution time threshold are discarded to ensure validity and computational efficiency.

\vspace{-1.0em}
\subsection{Hypothesis Optimization and Assessment}
\label{sec:method-eval}
\vspace{-0.8em}
After generating equation skeleton hypotheses, we evaluate and score them using observed data (Fig.~\ref{fig:llm-sr}(b)). This process involves optimizing the parameters of each hypothesis and then assessing its fitness. We decouple equation discovery into two steps: 
($i$)~discovering the equation program structures (skeletons) using the LLM, and ($ii$)~optimizing the skeleton parameters/coefficients based on data. The LLM is responsible for generating equation skeletons and the core logic of the program, while the numeric values of the parameters are represented as placeholders in the form of a parameter vector $\texttt{params}$ (as shown in Fig.~\ref{fig:spec-bacteria}). These placeholders are subsequently optimized to fit the data.
Each equation program skeleton $f \in \mathcal{F}_t$ is a function of the form: ``\lstinline[language=Python, style=inlinestyle]{def f(x, params): ... return y}''. We employ two optimization approaches: \textbf{numpy+BFGS}: A nonlinear optimization method using \texttt{scipy} library \citep{bfgs_flet87}, and \textbf{torch+Adam}: A stochastic gradient-based optimization algorithm using \texttt{PyTorch} \citep{adam-kingma}.
The choice between these methods depends on the problem characteristics and equation skeleton complexity. The numpy+BFGS is preferred for problems with fewer parameters, while torch+Adam is more suitable for larger-scale problems benefiting from efficient gradient computation through differentiable programming.

After optimizing the skeleton parameters ($\texttt{params}^*$), we assess the fitness of equation program hypotheses by measuring its ability to capture underlying patterns in the data.  We compute predicted target values as: $\hat{\mathbf{y}} = f(\mathbf{x}, \texttt{params}^*)$. The fitness evaluation score $s$ is then calculated as the negative Mean Squared Error (MSE) between predicted and true target values: $s = \text{Score}_\mathcal{T}(f, \mathcal{D}) = -\text{MSE}\left(\hat{\mathbf{y}},\mathbf{y}\right)$.


\vspace{-1.0em}
\subsection{Experience Management}
\label{sec:method-exp}
\vspace{-0.8em}
To efficiently navigate the search landscape and avoid local minima, \modelname employs an experience management step (Fig.\ref{fig:llm-sr}(c)). This process maintains a diverse population of high-quality equation programs in a dynamic experience buffer and samples from this population to construct informative prompts for subsequent LLM iterations.
Let $\mathcal{P}_t$ denote the experience buffer at iteration $t$, storing pairs of equation skeleton hypotheses and their corresponding scores $(f, s)$. We adopt an islands model \citep{pysr-2023,FunSearch} with $m$ independently evolving islands, initialized with a copy of the equation program example from the initial prompt (\texttt{equation$\_$v0} in Fig.~\ref{fig:spec-bacteria}). 
At each iteration $t$, new hypotheses $\mathcal{F}_t$ and their scores are added to the source island (from which the in-context examples of prompts were sampled) if they improve upon the current best: $\mathcal{P}_t^i \leftarrow \mathcal{P}_t^i \cup \{(f, s): f \in \mathcal{F}_t, s = -\text{Score}_{\mathcal{T}}(f, \mathcal{D}), s > s_{\text{best}}^i\}$ where $\mathcal{P}_t^i$ is the $i$-th island and $s_{\text{best}}^i$ is its current best score. Within each island, equation programs are clustered based on their signature (defined by their score) to further preserve diversity.

To construct informative prompts, we then sample equation programs from the experience buffer using a two-stage method. First, uniformly select a random island from the $m$ available. Second, sample $k$ equation programs from the selected island using
(a)~Cluster selection via Boltzmann sampling, favoring higher scores: $P_i = \frac{\exp\left(s_i / \tau_{c}\right)}{\sum_{i'} \exp\left(s_i' / \tau_{c}\right)}$ where $s_i$ is the mean score of the $i$-th cluster and $\tau_{c}$ is a temperature parameter. (b)~Individual program sampling, favoring shorter programs: $P(f_i) \propto \exp(-\tilde{l}_i / \tau_{p})$ where $\tilde{l}_i$ is the normalized program length and $\tau_p$ is a temperature parameter. The sampled programs are then included in the prompt as in-context experience demonstrations, guiding the LLM in generating new equation program hypotheses.
Detailed sampling procedures are provided in Appendix~\ref{sec:app-method}.

\begin{wrapfigure}[22]{r}{0.4\linewidth}
\vspace{-2.0em}
\centering
\begin{minipage}{\linewidth}
\begin{algorithm}[H]
\label{alg:llmsr}
\fontsize{8.5}{8.5}\selectfont
\SetAlgoLined
\SetKwInOut{Require}{Require}
\SetKwInOut{Input}{Input}
\SetKwInOut{Output}{Output}
\SetKw{KwDownTo}{downto}
\SetKw{KwAnd}{and}
\SetKw{KwContinue}{continue}
\caption{LLM-SR}
\Input{LLM $\pi_{\theta}$, dataset $\mathcal{D}$, problem $\mathcal{T}$, \\
$T$ iterations, $k$ in-context examples, \\$b$ samples per prompt}
\BlankLine
{\textcolor{gray}{\# Initialize population}}\\
{$\mathcal{P}_0 \gets \text{InitPop}()$}\\
$f^*, s^* \gets \text{null}, -\infty$\\
\For{$t \gets 1$ \KwTo $T-1$}{
{\textcolor{gray}{\# Sample examples from buffer}\\
$E \gets \{e_j\}_{j=1}^k$, $e_j = \text{SampleExp}(\mathcal{P}_{t-1})$}\\
{\textcolor{gray}{\# Prompt with new examples}\\
$\mathbf{p} \gets \text{MakeFewShotPrompt}(E)$}\\
{\textcolor{gray}{\# Sample from LLM}\\
$\mathcal{F}_t \gets \{f_j\}_{j=1}^b$, $f_j \sim \pi_\theta(\cdot | \mathbf{p})$}\\
{\textcolor{gray}{\# Evaluation and population update}}\\
\For{$f \in \mathcal{F}_t$}{
{$s \gets \text{Score}_{\mathcal{T}}(f, \mathcal{D})$}\\
\If{$s > s^*$}{
{$f^*, s^* \gets f, s$}\\
{$\mathcal{P}_{t} \gets \mathcal{P}_{t-1} \cup \{(f,s)\}$}
}
}
}
\BlankLine
\Output{$f^*, s^*$} 
\end{algorithm}
\end{minipage}
\end{wrapfigure}

Algorithm~\ref{alg:llmsr} presents the simplified pseudo-code of the \modelname framework. The experience buffer $\mathcal{P}_0$ is initialized with initial prompt, using a simple linear equation skeleton as a template (e.g., Fig.~\ref{fig:spec-bacteria} for the nonlinear oscillator problem). This initial structure serves as a baseline for the LLM to modify operators and structures based on its domain knowledge. Each iteration $t$ involves: ($i$)~sampling $k$ in-context examples from $\mathcal{P}_{t-1}$, ($ii$)~updating the prompt, ($iii$)~generating $b$ equation program skeletons from the LLM, and ($iv$)~evaluating and potentially adding these to $\mathcal{P}_t$ if they improve upon the best score $s^*$. 
This process leverages the LLM's generative capabilities to refine equation structures guided by the evolving experience buffer. The algorithm returns the best-scoring program $f^*$ and its score $s^*$ as the optimal solution, iteratively exploring the equation space while balancing exploitation of promising structures with exploration of new possibilities.

\vspace{-1.0em}
\section{Experiments}
\label{sec:exp-setup}
\vspace{-0.8em}
\subsection{Benchmarks and Datasets}
\vspace{-0.8em}
\label{sec:exp-setup-data}
The Feynman benchmark \citep{AI-Feynman-Science-2020}, comprising 120 fundamental physics problems from \textit{Feynman
Lectures on Physics database series}\footnote{\url{https://space.mit.edu/home/tegmark/aifeynman.html}}, is the current standard benchmark for evaluating symbolic regression techniques in scientific equation discovery. However, our investigation reveals that LLMs have significant memorization issues with these well-known physics equations, potentially undermining their effectiveness in assessing LLM-based equation discovery approaches. 
For instance, \modelname rapidly achieves low data-driven errors within few iterations ($<20$) on Feynman problems, suggesting a recitation of memorized information rather than a process of discovery (full results in App.~\ref{sec:app-feyn}).
To address these limitations and provide a more robust evaluation, we introduce novel benchmark problems across three scientific domains. Our benchmark design focuses on: ($i$)~Custom modifications to physical models to prevent trivial memorization; ($ii$)~Complex synthetic non-linear relationships to challenge creative exploration; and ($iii$)~Realistic scenarios with experimental data to reflect real modeling processes. These benchmarks are designed to challenge the model's ability to uncover complex mathematical relations while leveraging its scientific prior knowledge, simulating conditions for scientific discovery. 
To validate our new benchmarks' effectiveness, we compared LLM response perplexity (using open-source Mixtral-8x7B) and equation discovery error curves (using GPT-3.5) between Feynman problems and our new benchmarks. Results show lower perplexity (Fig.~\ref{fig:feyn_perplexity} in App.~\ref{sec:app-feyn}) and sharper discovery curves (Fig.~\ref{fig:feyn-all} in App.~\ref{sec:app-feyn}) for Feynman problems, suggesting that both LLM backbones have more likely memorized common Feynman equations, while our benchmarks present novel challenges requiring reasoning and exploration.
We next discuss these new benchmark problems in detail:

\vspace{-0.8em}
\paragraph{Nonlinear Oscillators}
Nonlinear damped oscillators, ubiquitous in physics and engineering, are governed by differential equations describing the complex interplay between an oscillator's position, velocity, and acting forces. The general form of these equations is typically expressed as: $\ddot{x} + f(t, x, \dot{x}) = 0$ where $t$ is time, $x$ is position, and $f(t,x, \dot{x})$ represents nonlinear forces. To challenge LLM-based equation discovery methods beyond common oscillator systems (e.g., Van der Pol, Duffing), we introduce two custom nonlinear designs: \textcircled{1}~\texttt{Oscillation~1}:~$
\dot{v}=~F\sin(\omega x)-\alpha v^3-\beta x^3-\gamma xv-x\cos(x)
$; and \textcircled{2}~\texttt{Oscillation~2}:~$
\dot{v}=F\sin(\omega t)-\alpha v^3-\beta xv -\delta x\exp(\gamma x)$, where $v=\dot{x}$ represents velocity, and $\omega, \alpha, \beta, \gamma, \delta$ are constants.
These two forms, serving as a proof of concept, are carefully designed to incorporate a combination of challenging yet solvable nonlinear structures (including trigonometric, polynomial, and exponential) that are distinct from well-known oscillator systems.
More details on the design rationale and data generation are provided in App.~\ref{sec:app-oscillator}.

\vspace{-0.8em}
\paragraph{Bacterial Growth}
The growth of Escherichia coli (E. coli) bacteria has been widely studied in microbiology due to its importance in various applications, such as biotechnology, and food safety. Discovering equations governing E. coli growth rate under different conditions is crucial for predicting and optimizing bacterial growth.
The bacterial population growth rate has been modeled using a differential equation with the effects of population density~($B$), substrate concentration~($S$), temperature~($T$), and $\text{pH}$ level, which is commonly formulated with multiplicative structure: $\frac{dB}{dt} = f(B,S,T,\text{pH}) = f_B(B) \cdot f_S(S) \cdot f_T(T) \cdot f_{\text{pH}}(\text{pH})$.
To create a challenging benchmark that leverages LLMs' prior knowledge while preventing trivial memorization, we introduce novel nonlinear formulations for $f_T(T)$ and $f_{\text{pH}}(\text{pH})$.
These custom functions maintain key characteristics of established models while introducing complexities that require exploration and discovery rather than recall.
The complete mathematical formulations, along with the data generation process and parameter ranges, are detailed in App.~\ref{sec:app-bacterial}.

\vspace{-0.8em}
\paragraph{Material Stress Behavior}
The stress-strain relationship of materials under varying conditions, particularly as a function of temperature and material type, is fundamental to structural design and analysis across engineering disciplines. This benchmark problem leverages a real-world experimental dataset from \citep{stress-data-brief}, comprising tensile tests on Aluminum 6061-T651 across a range of temperatures. The inclusion of this benchmark serves multiple purposes: ($i$)~It challenges LLM-based equation discovery methods with experimental data, moving beyond synthetic or idealized problems. ($ii$)~Unlike the previous benchmarks, there is no predetermined theoretical model structure for this problem, necessitating creative modeling approaches from LLMs. In other words, modeling for this type of task is mostly empirical and the stress-strain-temperature relations may vary significantly based on the specific material and experimental condition, preventing trivial memorization.
More details on this problem and experimental data are provided in App.~\ref{sec:app-stress}.



\vspace{-1.0em}
\subsection{Experimental Setup}
\label{sec:exp-setup-baseline}
\vspace{-0.8em}
We compare \modelname against state-of-the-art symbolic regression (SR) methods, including evolutionary-based approaches like \textbf{GPlearn}\footnote{\url{https://gplearn.readthedocs.io/en/stable/}} (Genetic Programming) and \textbf{PySR}\footnote{\url{https://github.com/MilesCranmer/PySR}} (multi-island asynchronous evolution) \citep{pysr-2023}, and deep learning-based methods such as \textbf{DSR} (reinforcement learning for expression generation) \citep{DSR-Petersen-ICLR-2021} and \textbf{uDSR} (extending DSR with Genetic Programming search at decoding) \citep{UDSR-NeurIPS-2022}. 
We also evaluate against pre-trained Transformer SR models: \textbf{NeSymReS} \citep{Biggio-NeSymReS-ICML-2021} and \textbf{E2E} \citep{Kamienny-E2E-symbolic-NIPS-2022}.
This selection provides a comprehensive evaluation across different SR paradigms. 
We allow all search-based baselines to run for over $2$M iterations until convergence to their best performance.
In \modelname experiments, each iteration samples $b=4$ equation skeletons per prompt with temperature $\tau=0.8$, optimizes parameters via numpy+BFGS or torch+Adam (with $30$ seconds timeout), and uses $k=2$ in-context examples from the experience buffer for refinement. We run \modelname variants for around $2.5$K iterations in all experiments. 
More details on the implementation and parameter settings of each baseline as well as implementation specifics of \modelname, including experience buffer structure, prompt refinement strategy, and parallel evaluation are provided in App.~\ref{sec:app-imp}.
\label{sec:exp-findings}
\vspace{-0.8em}
\subsection{Quantitative Results}
\vspace{-0.8em}
\paragraph{Accuracy}
Table~\ref{table-result} compares the performance of \modelname (using GPT-3.5 and Mixtral backbones) against state-of-the-art symbolic regression methods across various scientific benchmarks. Performance is measured using Normalized Mean Squared Error (NMSE), with lower values indicating better performance.
\modelname with both backbones consistently outperform baselines, despite running for fewer iterations ($2.5$K vs. $2$M+ for baselines). 
To assess generalization capability, we evaluate performance on both in-domain (ID) and out-of-domain (OOD) test sets. 
The performance gap between \modelname and baselines is more pronounced in the OOD setting, suggesting superior generalization of \modelname's discovered equations. 
For instance, on the E. coli growth problem, \modelname achieves an OOD NMSE of $\sim 0.0037$, significantly outperforming other methods (with OOD NMSE $>1$).

\begin{wrapfigure}[22]{r}{0.57\linewidth}
\centering
\vspace{-1.2em}
\includegraphics[width=\linewidth]{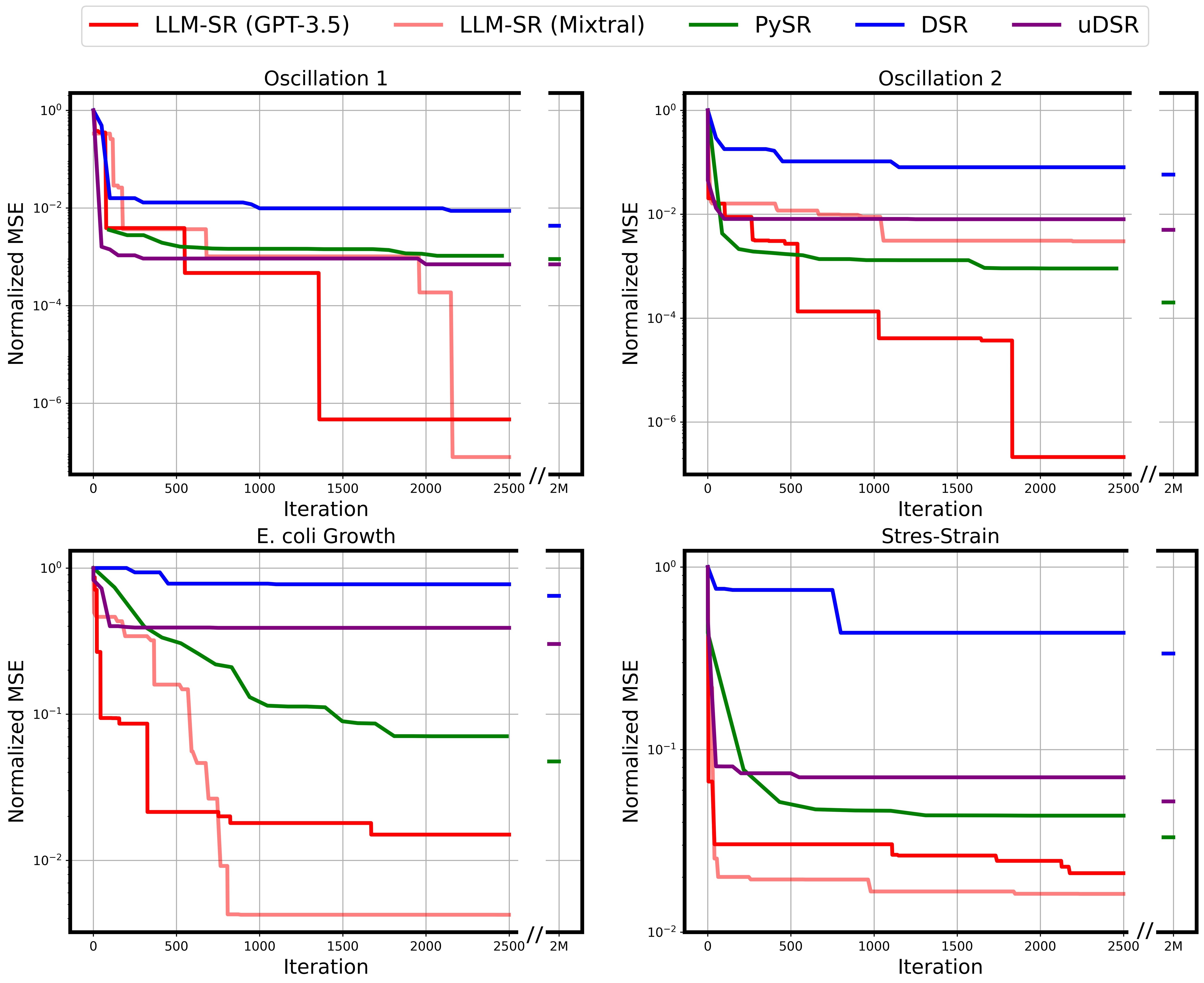}
\caption{
\small 
\textbf{Best score trajectories} of \modelname with GPT-3.5 and Mixtral against SR baselines across different benchmark problems.
\modelname discovers accurate equations more efficiently, requiring fewer iterations.
Baselines fail to match \modelname even after $2$M iterations.
}
\vspace{-0.5em}
\label{fig:loss-curve-all}
\vspace{-0.5em}
\end{wrapfigure}
Among baselines, PySR and uDSR show the best performance, while Transformer SR models (NeSymReS, E2E) perform poorly, likely due to limited generalization from their pretraining on common benchmark distributions to our novel datasets.
These results demonstrate \modelname's effectiveness in discovering accurate and generalizable equations across diverse scientific domains.

\begin{table}[t]
\centering
\resizebox{0.9\columnwidth}{!}{
\begin{tabular}{lcccccccc
}
\toprule
\multirow{2}{*}{Model} & \multicolumn{2}{c}{Oscillation 1} & \multicolumn{2}{c}{Oscillation 2} & \multicolumn{2}{c}{E. coli growth} & \multicolumn{2}{c}{Stress-Strain}  \\
\cmidrule(lr){2-3} \cmidrule(lr){4-5} \cmidrule(lr){6-7} \cmidrule(lr){8-9}
 & ID$\downarrow$   & OOD$\downarrow$ & ID$\downarrow$ & OOD$\downarrow$ & ID$\downarrow$ & OOD$\downarrow$  & ID$\downarrow$  & OOD$\downarrow$ 
 \\
\midrule
GPlearn & 0.0155 & 0.5567 & 0.7551 & 3.188 & 1.081 & 1.039 & 0.1063 & 0.4091 \\
NeSymReS \citep{Biggio-NeSymReS-ICML-2021} & 0.0047 & 0.5377 & 0.2488 & 0.6472 & \multicolumn{2}{c}{\textcolor{gray}{N/A ($d>3$)}} & 0.7928& 0.6377 \\
E2E \citep{Kamienny-E2E-symbolic-NIPS-2022} & 0.0082 & 0.3722 & 0.1401 & 0.1911 & 0.6321 & 1.4467 & 0.2262 & 0.5867 \\
DSR \citep{DSR-Petersen-ICLR-2021} & 0.0087 & 0.2454 & 0.0580 & 0.1945 & 0.9451 & 2.4291 & 0.3326 & 1.108 \\
uDSR \citep{UDSR-NeurIPS-2022} & \underline{0.0003} & \underline{0.0007} & 0.0032 & \underline{0.0015} & 0.3322 & 5.4584 & 0.0502 & 0.1761 \\
PySR \citep{pysr-2023} & 0.0009 & 0.3106 & \underline{0.0002} & 0.0098 & \underline{0.0376} & \underline{1.0141} & \underline{0.0331} & \underline{0.1304} \\
\midrule
\modelname (\texttt{Mixtral}) &\cellcolor{mygray}\textbf{7.89e-8}& \cellcolor{mygray}\textbf{0.0002} & 0.0030 & 0.0291 & \cellcolor{mygray}\textbf{0.0026} & \cellcolor{mygray}\textbf{0.0037} & \cellcolor{mygray}\textbf{0.0162} & 0.0946 \\
\modelname (\texttt{GPT-3.5}) & 4.65e-7 & 0.0005 & \cellcolor{mygray}\textbf{2.12e-7} & \cellcolor{mygray}\textbf{3.81e-5} & 0.0214 & 0.0264 & 0.0210 & \cellcolor{mygray}\textbf{0.0516} \\
\arrayrulecolor{black}\bottomrule
\arrayrulecolor{black}\bottomrule
\end{tabular}}
\vspace{-0.5em}
\caption{\small \textbf{Quantitative performance comparison} of \modelname (with 
GPT-3.5 and Mixtral backbones), and SR baseline models
on different scientific benchmark problems measured by Normalized Mean Squared Error.
``N/A'' refers to incompatibility of E. coli Growth dataset for the NeSymReS baseline (limited to $<3d$ data). 
\vspace{-0.5em}
}
\vspace{-0.5em}
\label{table-result}
\end{table}

\vspace{-0.7em}
\paragraph{Efficiency}
Fig.~\ref{fig:loss-curve-all} shows the performance trajectories of \modelname variants and symbolic regression baselines across different scientific benchmark problems, depicting the best fitting scores achieved over search iterations. 
By leveraging scientific prior knowledge, \modelname explores a considerably lower number of equation candidates in the vast optimization space compared to symbolic regression baselines that lack this knowledge. This is evident from the sharp drops in the error curves for \modelname variants, indicating they efficiently navigate the search space by exploiting domain knowledge to identify promising candidates more quickly. 
In contrast, the symbolic regression baselines show much more gradual improvements and fail to match \modelname's performance even after $2$M+ iterations.
The performance gap between \modelname and baselines also mostly widens over iterations, highlighting the effectiveness of LLMs acting as mutation (or crossover) agents in \modelname's iterative refinement process.

\vspace{-0.5em}
\subsection{Qualitative Analysis}
\vspace{-0.5em}

\begin{figure}[t]
\centering
\includegraphics[width=\linewidth]{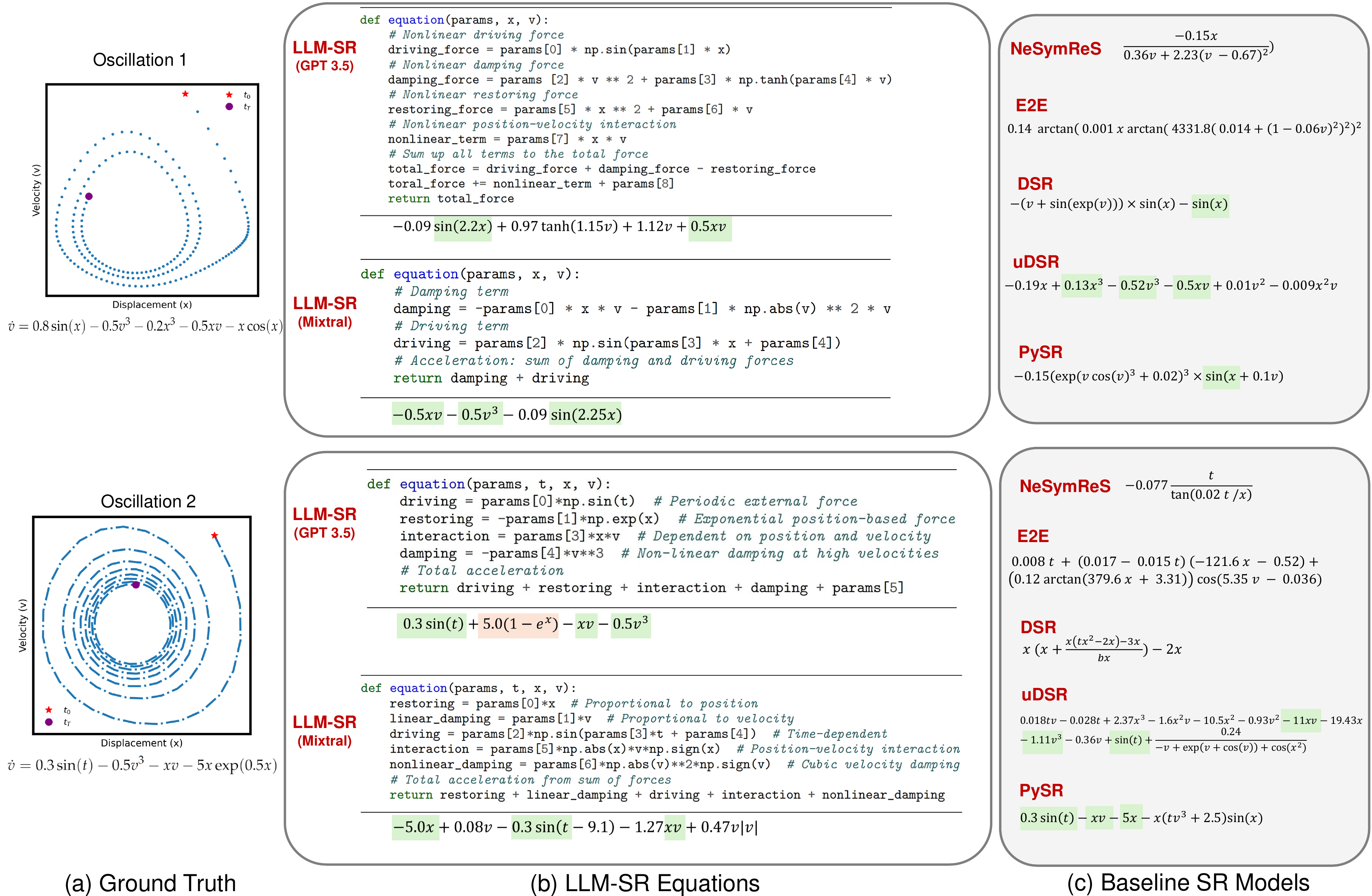}
\caption{\small \textbf{Discovered equations for Oscillation 1 (top) and Oscillation 2 (bottom) problems}: \textbf{(a)}~True equations and their phase diagram; \textbf{(b)}~Equation program skeletons identified by \modelname, with simplified forms obtained after parameter optimization; and \textbf{(c)}~Equations found using SR baselines. Shaded green terms denote recovered symbolic terms from true equations. 
\vspace{-0.5em}
}
\label{fig:osc_equations}
\vspace{-0.5em}
\end{figure}
\paragraph{Discovered Equations}
Fig.~\ref{fig:osc_equations} presents the final discovered equations for both Oscillation problems using \modelname and other symbolic regression baselines. A notable observation is that equations discovered by \modelname have better recovered the symbolic terms of the true equations compared to baselines. Also, \modelname provides explanations and reasoning steps based on scientific knowledge about the problem, leading to more interpretable terms combined as the final function.
For example, in both problems, \modelname identifies the equation structure as a combination of driving force, damping force, and restoring force terms, relating them to the problem's physical characteristics. 
In contrast, baselines generate equations lacking interpretability and understanding of the physical meanings of variables and the relations between them.
These equations appear as a combination of mathematical operations and variables without a clear connection to the problem's underlying physical principles.
App.~\ref{sec:app-results} provides a 
more detailed qualitative analysis of the final discovered equations for other benchmark problems (Figs.~\ref{fig:bacterial_equations} and \ref{fig:stress_equations}), as well as the equations discovered over the performance trajectory of \modelname's iterations (Figs.~\ref{fig:osc2-annotated}-\ref{fig:ss-annotated}).

\begin{wrapfigure}[14]{r}{0.5\linewidth}
\centering
\vspace{-1.0em}
\includegraphics[width=\linewidth]{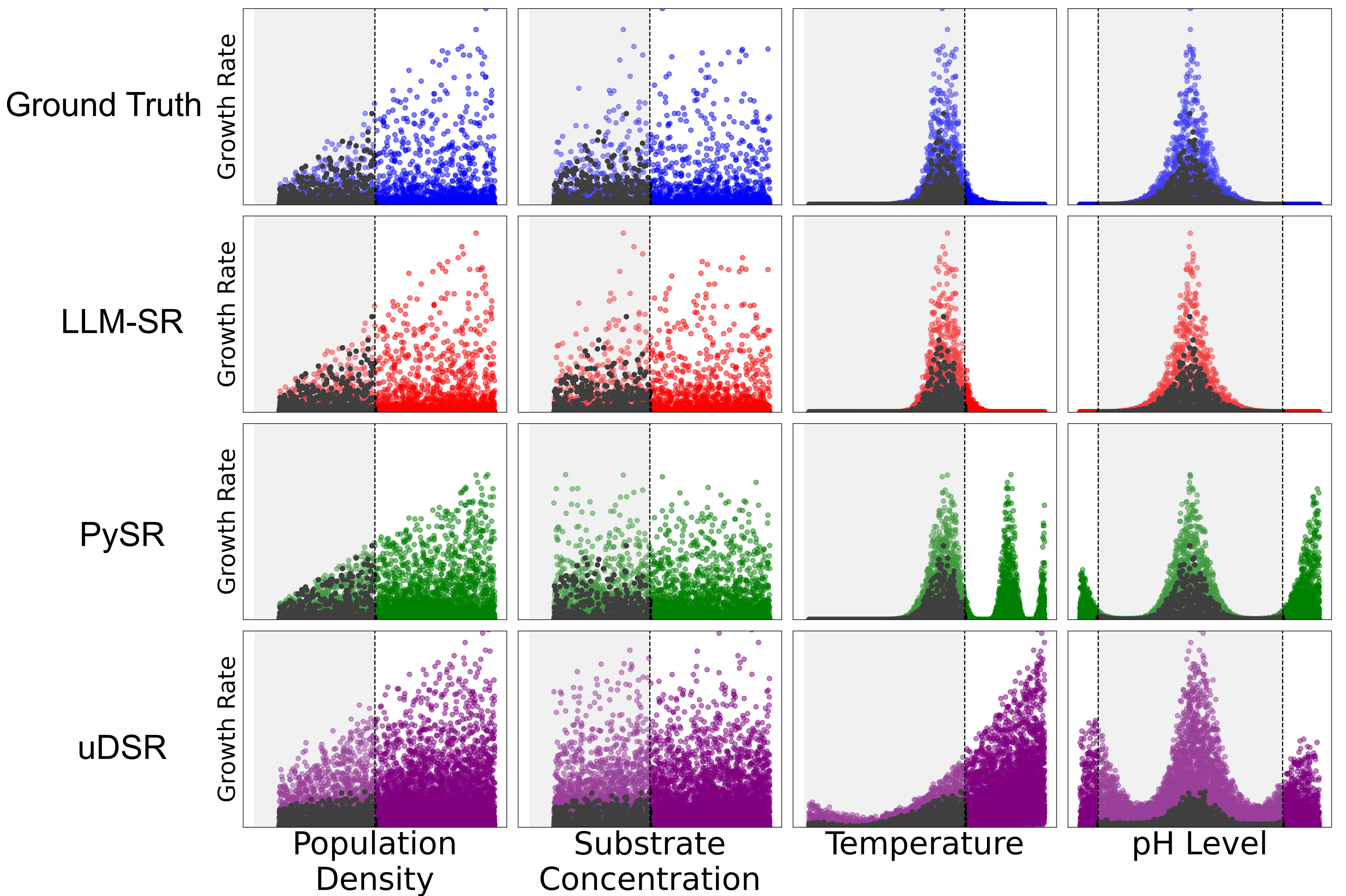}
\caption{\small 
Comparison of E. coli growth rate distributions from \modelname, PySR, and uDSR. 
}
\label{fig:qual_bacterial}
\end{wrapfigure}

\vspace{-0.8em}
\paragraph{Generalization}
Fig.~\ref{fig:qual_bacterial} compares predicted distributions obtained from \modelname, and competing baselines (PySR and uDSR) with the ground truth distribution of E. coli growth problem. 
The shaded region and black points indicate in-domain (ID) data, while the rest represent out-of-domain (OOD). 
Results show that distributions obtained from \modelname align well with the ground truth, not only for ID data but also for OOD regions. This alignment demonstrates the better generalizability of equations discovered by \modelname to unseen data, likely due to the integration of scientific prior knowledge in the equation discovery process.
In contrast, PySR and uDSR tend to overfit the observed data, with significant deviations in OOD regions. 

This overfitting behavior highlights their limited ability to generalize beyond the training data and capture the true physical underlying patterns.
Detailed analyses for other benchmark problems are provided in App.~\ref{sec:app-results}.

\vspace{-0.5em}
\subsection{Ablation Study}
\label{sec:exp-analysis-ab}
\begin{figure}[h]
\centering
\centering
\includegraphics[width=0.7\linewidth]{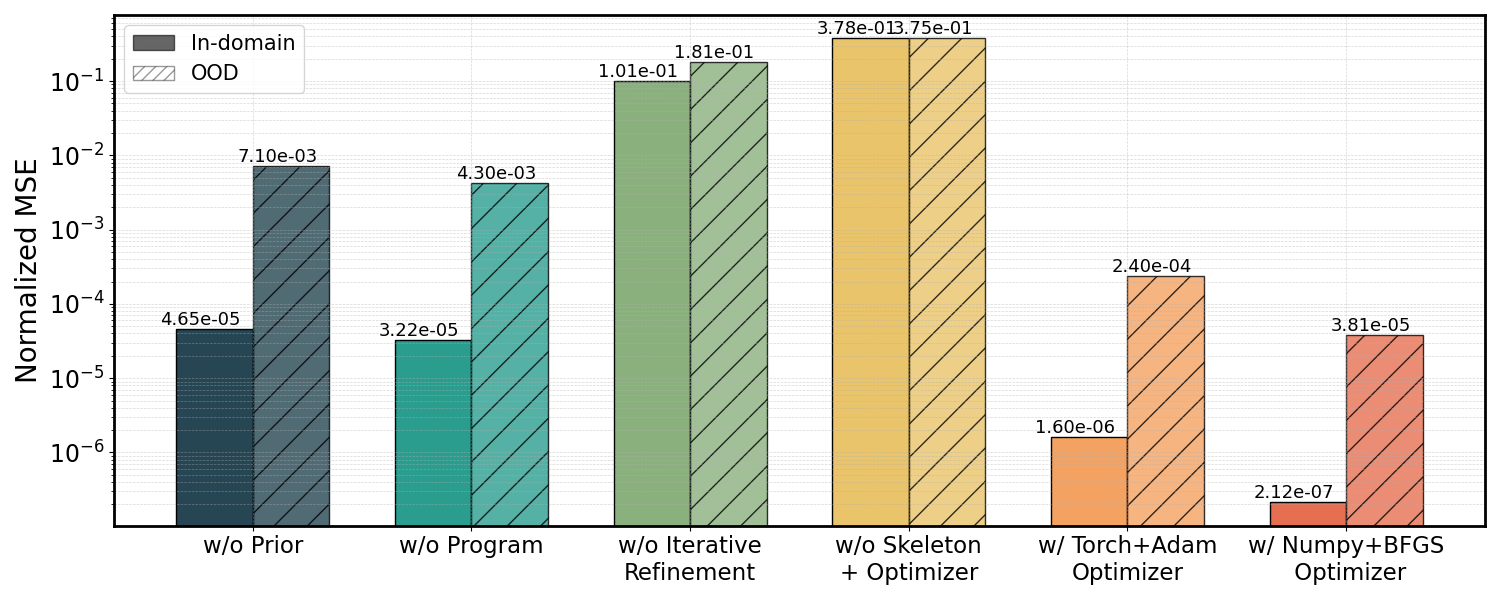}
\caption{\small 
\textbf{Ablation results on the Oscillation~2 problem}, showing the impact of problem specification, program representation, iterative refinement, parameter optimization, and optimization frameworks on  \modelname's performance.}
\label{fig:ablation}
\end{figure}

\vspace{-0.5em}
We conducted an ablation study on the Oscillation 2 problem using GPT-3.5 as the LLM backbone to investigate the impact of \modelname's key components (Fig.~\ref{fig:ablation}, more detailed results in App.~\ref{sec:app-ablation}). Our findings reveal the crucial role of each component in the model's performance.
The ``\textit{\textbf{w/o Prior}}'' variant, which removes the natural language description of the scientific problem and its variables, led to a considerable performance drop. This highlights the importance of incorporating prior domain knowledge in equation discovery.
The ``\textit{\textbf{w/o Program}}'' variant, which 
restricts LLM hypothesis generation to single-line mathematical expressions, also had a negative but less severe impact on performance,
denoting the importance of programming flexibility in this task.
The ``\textit{\textbf{w/o Iterative Refinement}}'' variant, equivalent to the LLM sampling without the optimization loop, led to substantial performance drops (NMSE: 1.01e-1 in-domain, 1.81e-1 OOD), emphasizing the importance of the evolutionary search and optimization process in \modelname's success. The ``\textit{\textbf{w/o skeleton + optimizer}}''
variant, which requires end-to-end equation generation without separate parameter optimization step (i.e., generating hypotheses as full equations along with their numeric parameters), also significantly worsened results (NMSE: 3.78e-1 in-domain, 3.75e-1 OOD). This highlights the effectiveness of our two-stage approach—generating equation skeletons followed by data-driven parameter optimization—in navigating complex combinatorial optimization space of discrete equation structures and continuous parameters.

We compared two optimization frameworks: \textbf{numpy+BFGS} and \textbf{torch+Adam}. In our experiments, the numpy+BFGS variant performed slightly better 
compared to torch+Adam. 
This difference is most likely attributed to the LLM's higher proficiency in generating numpy code rather than inherent superiority of the optimization method for this task. 
\modelname relies on direct and differentiable parameter optimization, a capability not present in current symbolic regression methods.
Combining \modelname with LLM backbones that are better in generating PyTorch code could potentially enhance equation discovery by leveraging differentiable parameter optimization in future.

\section{Related Work}
\label{sec:related}
\vspace{-0.5em}
\paragraph{LLMs and Optimization}

While LLMs have shown remarkable capabilities in various domains, their performance often falls short in tasks requiring high precision and complex reasoning. To address this, researchers have explored combining LLMs with feedback mechanisms \citep{self_refine-2024, LeanDojo-theorem, llm-program-self} and integrating them into iterative optimization loops \citep{Lehman-ELM, llambo-llm-bayesian, LLM-Evo-Survey, LLM-Evo-Lange-2024}. Recently, LLMs have been successfully applied in prompt optimization \citep{LLM-Optimizer, connect-LLM-Evo-prompt}, data-driven analysis \citep{data-driven-AI2-LM, LLM-scientific-synthesis}, and neural architecture search \citep{EvoPrompt-2023, gpt4-NAS-llm}.
Most related to our work is FunSearch \citep{FunSearch} that combines LLMs with systematic evaluators to search for heuristics that push the boundaries in solving some established open mathematical problems.
Building upon these ideas, our \modelname framework employs LLM as an optimizer, leveraging its scientific prior knowledge and data-driven evaluators to discover mathematical equations underlying scientific observations.

\vspace{-0.8em}
\paragraph{LLMs for Scientific Discovery}
The integration of LLMs into scientific tasks has recently garnered significant attention, offering transformative potential across various fields such as drug discovery, biology, and materials science \citep{scientific-age-AI, llm-scientific-gpt-microsoft}. 
Specifically, recent studies have demonstrated the capacity of LLMs to propose scientifically plausible and potentially novel hypotheses by leveraging their extensive domain knowledge and reasoning capabilities \citep{data-driven-AI2-LM, LLM-scientific-synthesis,LLM-scientific-zero-shot-hypothesis, ji2024scimon}. Also, when equipped with external tools and scientific simulators, LLM agents have shown promise in automated statistical discovery and reasoning \citep{li2024automated, wang2023hypothesis, ma2024llm-bilevel}.
Despite the increasing exploration of LLMs in scientific contexts and question answering, their potential for tasks such as equation discovery and symbolic regression remains largely unexplored.
Our work extends this line of research by introducing a novel approach for equation discovery that combines LLMs' scientific prior knowledge and code generation with data-driven evaluation.


\vspace{-0.8em}
\paragraph{Symbolic Regression}
Symbolic regression (SR) methods can be broadly categorized into search-based approaches, learning-based models, as well as hybrid learning and search methods. Search-based approaches mainly explore the space of equation structures and parameters using evolutionary algorithms or reinforcement learning \citep{Schmidt-Lipson-2009, pysr-2023, DSR-Petersen-ICLR-2021, SPL-MCTS-ICLR-2023}. They offer interpretable results but often struggle with scalability and efficiency.
Learning-based models, on the other hand, leverage large-scale synthetic data and Transformer models to learn the mapping between numeric input observations and output mathematical expressions \citep{Biggio-NeSymReS-ICML-2021, Kamienny-E2E-symbolic-NIPS-2022}. Hybrid methods aim to combine the strengths of both approaches, guiding the search by employing neural priors to improve the expressiveness and efficiency of the discovery process \citep{UDSR-NeurIPS-2022, TPSR-NeurIPS, mundhenk-seeding-GP-NeurIPS-2021, SNIP-ICLR}. Despite the progress made by these approaches, they often face limitations such as the lack of scientific prior knowledge incorporation and the restricted expressiveness of traditional equation representations like expression trees.
While there have been some works 
incorporating prior knowledge
by using declarative bias and structures with pre-defined grammars
 \citep{Todorovski-Declarative-SR, Todorovski-integrate-SR},
 these methods do not leverage the power of LLMs for this task. Our work advances this research direction by utilizing LLMs to efficiently search the combinatorial optimization space of equation discovery and generate meaningful equation structures based on the embedded scientific prior knowledge.

\section{Conclusion and Future Work}
\label{sec:discussion}
\vspace{-.5em}
In this work, we introduced \modelname, a novel approach to equation discovery that leverages the scientific knowledge and code generation capabilities of Large Language Models (LLMs). By treating equations as programs and combining LLM-generated educated hypotheses with evolutionary search, our method demonstrates superior performance on benchmark problems across diverse scientific domains, particularly in out-of-domain test settings.
Despite its promising results, \modelname has limitations. The method's performance is inherently tied to the quality and breadth of the LLM's training data, which may lead to biases or gaps in certain scientific domains. Additionally, the computational cost of iterative LLM queries and parameter optimization could be prohibitive for large-scale problems.
Future work could focus on integrating domain-specific LMs and retrieval-augmented learning techniques to enhance the relevance and accuracy of generated equations; and incorporating human domain experts in the pipeline to improve the scientific plausibility. 
The creation of more comprehensive benchmarks, designed to simulate true discovery processes and prevent LLM recitation, is also crucial for rigorous evaluation of LLM-based equation discovery methods.

\subsection*{Acknowledgments}
This research was partially supported by the U.S. National Science Foundation (NSF) under Grant No. 2416728.

\newpage


\bibliography{main}

\begin{thebibliography}{50}
\providecommand{\natexlab}[1]{#1}
\providecommand{\url}[1]{\texttt{#1}}
\expandafter\ifx\csname urlstyle\endcsname\relax
  \providecommand{\doi}[1]{doi: #1}\else
  \providecommand{\doi}{doi: \begingroup \urlstyle{rm}\Url}\fi

\bibitem[Aakash et~al.(2019)Aakash, Connors, and Shields]{stress-data-brief}
B.S. Aakash, JohnPatrick Connors, and Michael~D. Shields.
\newblock Stress-strain data for aluminum 6061-t651 from 9 lots at 6 temperatures under uniaxial and plane strain tension.
\newblock \emph{Data in Brief}, 25:\penalty0 104085, 2019.
\newblock ISSN 2352-3409.
\newblock \doi{https://doi.org/10.1016/j.dib.2019.104085}.

\bibitem[Achiam et~al.(2023)Achiam, Adler, Agarwal, Ahmad, Akkaya, Aleman, Almeida, Altenschmidt, Altman, Anadkat, et~al.]{achiam2023gpt}
Josh Achiam, Steven Adler, Sandhini Agarwal, Lama Ahmad, Ilge Akkaya, Florencia~Leoni Aleman, Diogo Almeida, Janko Altenschmidt, Sam Altman, Shyamal Anadkat, et~al.
\newblock Gpt-4 technical report.
\newblock \emph{arXiv preprint arXiv:2303.08774}, 2023.

\bibitem[AI4Science \& Quantum(2023)AI4Science and Quantum]{llm-scientific-gpt-microsoft}
Microsoft~Research AI4Science and Microsoft~Azure Quantum.
\newblock The impact of large language models on scientific discovery: a preliminary study using gpt-4.
\newblock \emph{arXiv preprint arXiv:2311.07361}, 2023.

\bibitem[Biggio et~al.(2021)Biggio, Bendinelli, Neitz, Lucchi, and Parascandolo]{Biggio-NeSymReS-ICML-2021}
Luca Biggio, Tommaso Bendinelli, Alexander Neitz, Aurelien Lucchi, and Giambattista Parascandolo.
\newblock Neural symbolic regression that scales.
\newblock In Marina Meila and Tong Zhang (eds.), \emph{Proceedings of the 38th International Conference on Machine Learning}, volume 139 of \emph{Proceedings of Machine Learning Research}, pp.\  936--945. PMLR, 18--24 Jul 2021.

\bibitem[Brence et~al.(2021)Brence, Todorovski, and Džeroski]{Todorovski-Prob-grammar-SR}
Jure Brence, Ljupčo Todorovski, and Sašo Džeroski.
\newblock Probabilistic grammars for equation discovery.
\newblock \emph{Knowledge-Based Systems}, 224:\penalty0 107077, 2021.
\newblock ISSN 0950-7051.
\newblock \doi{https://doi.org/10.1016/j.knosys.2021.107077}.

\bibitem[Brown(2020)]{brown2020language}
Tom~B Brown.
\newblock Language models are few-shot learners.
\newblock \emph{arXiv preprint arXiv:2005.14165}, 2020.

\bibitem[Chen et~al.(2023)Chen, Dohan, and So]{EvoPrompt-2023}
Angelica Chen, David Dohan, and David So.
\newblock Evoprompting: Language models for code-level neural architecture search.
\newblock In A.~Oh, T.~Neumann, A.~Globerson, K.~Saenko, M.~Hardt, and S.~Levine (eds.), \emph{Advances in Neural Information Processing Systems}, volume~36, pp.\  7787--7817. Curran Associates, Inc., 2023.

\bibitem[Cranmer(2023)]{pysr-2023}
Miles Cranmer.
\newblock Interpretable machine learning for science with pysr and symbolicregression. jl.
\newblock \emph{arXiv preprint arXiv:2305.01582}, 2023.

\bibitem[Fletcher(1987)]{bfgs_flet87}
Roger Fletcher.
\newblock \emph{Practical Methods of Optimization}.
\newblock John Wiley \& Sons, New York, NY, USA, second edition, 1987.

\bibitem[Guo et~al.(2024)Guo, Wang, Guo, Li, Song, Tan, Liu, Bian, and Yang]{connect-LLM-Evo-prompt}
Qingyan Guo, Rui Wang, Junliang Guo, Bei Li, Kaitao Song, Xu~Tan, Guoqing Liu, Jiang Bian, and Yujiu Yang.
\newblock Connecting large language models with evolutionary algorithms yields powerful prompt optimizers.
\newblock In \emph{The Twelfth International Conference on Learning Representations}, 2024.

\bibitem[Haluptzok et~al.(2022)Haluptzok, Bowers, and Kalai]{llm-program-self}
Patrick Haluptzok, Matthew Bowers, and Adam~Tauman Kalai.
\newblock Language models can teach themselves to program better.
\newblock \emph{arXiv preprint arXiv:2207.14502}, 2022.

\bibitem[Ji et~al.(2024)Ji, Wang, Downey, and Hope]{ji2024scimon}
Heng Ji, Qingyun Wang, Doug Downey, and Tom Hope.
\newblock Scimon: Scientific inspiration machines optimized for novelty.
\newblock In \emph{ACL Anthology: Proceedings of the 62nd Annual Meeting of the Association for Computational Linguistics (Volume 1: Long Papers), pages 279--299}. University of Illinois Urbana-Champaign/CABBI, 2024.

\bibitem[Jiang et~al.(2024)Jiang, Sablayrolles, Roux, Mensch, Savary, Bamford, Chaplot, Casas, Hanna, Bressand, et~al.]{mixtral-llm}
Albert~Q Jiang, Alexandre Sablayrolles, Antoine Roux, Arthur Mensch, Blanche Savary, Chris Bamford, Devendra~Singh Chaplot, Diego de~las Casas, Emma~Bou Hanna, Florian Bressand, et~al.
\newblock Mixtral of experts.
\newblock \emph{arXiv preprint arXiv:2401.04088}, 2024.

\bibitem[Kamienny et~al.(2022)Kamienny, d'Ascoli, Lample, and Charton]{Kamienny-E2E-symbolic-NIPS-2022}
Pierre-Alexandre Kamienny, St{\'e}phane d'Ascoli, Guillaume Lample, and Francois Charton.
\newblock End-to-end symbolic regression with transformers.
\newblock In \emph{Advances in Neural Information Processing Systems}, 2022.

\bibitem[Kingma \& Ba(2014)Kingma and Ba]{adam-kingma}
Diederik~P Kingma and Jimmy Ba.
\newblock Adam: A method for stochastic optimization.
\newblock \emph{arXiv preprint arXiv:1412.6980}, 2014.

\bibitem[Koza(1994)]{Koza1994-GP}
John~R. Koza.
\newblock Genetic programming as a means for programming computers by natural selection.
\newblock \emph{Statistics and Computing}, 4\penalty0 (2):\penalty0 87--112, Jun 1994.
\newblock ISSN 1573-1375.
\newblock \doi{10.1007/BF00175355}.

\bibitem[La~Cava et~al.(2021)La~Cava, Orzechowski, Burlacu, de~Franca, Virgolin, Jin, Kommenda, and Moore]{SRBench-Cava-NeurIPS-2021}
William La~Cava, Patryk Orzechowski, Bogdan Burlacu, Fabricio de~Franca, Marco Virgolin, Ying Jin, Michael Kommenda, and Jason Moore.
\newblock Contemporary symbolic regression methods and their relative performance.
\newblock In J.~Vanschoren and S.~Yeung (eds.), \emph{Proceedings of the Neural Information Processing Systems Track on Datasets and Benchmarks}, volume~1, 2021.

\bibitem[Landajuela et~al.(2022)Landajuela, Lee, Yang, Glatt, Santiago, Aravena, Mundhenk, Mulcahy, and Petersen]{UDSR-NeurIPS-2022}
Mikel Landajuela, Chak Lee, Jiachen Yang, Ruben Glatt, Claudio~P. Santiago, Ignacio Aravena, Terrell~N. Mundhenk, Garrett Mulcahy, and Brenden~K. Petersen.
\newblock A unified framework for deep symbolic regression.
\newblock In Alice~H. Oh, Alekh Agarwal, Danielle Belgrave, and Kyunghyun Cho (eds.), \emph{Advances in Neural Information Processing Systems}, 2022.

\bibitem[Lange et~al.(2024)Lange, Tian, and Tang]{LLM-Evo-Lange-2024}
Robert~Tjarko Lange, Yingtao Tian, and Yujin Tang.
\newblock Large language models as evolution strategies.
\newblock \emph{arXiv preprint arXiv:2402.18381}, 2024.

\bibitem[Langley(1981)]{langley1981data}
Pat Langley.
\newblock Data-driven discovery of physical laws.
\newblock \emph{Cognitive Science}, 5\penalty0 (1):\penalty0 31--54, 1981.

\bibitem[Lehman et~al.(2023)Lehman, Gordon, Jain, Ndousse, Yeh, and Stanley]{Lehman-ELM}
Joel Lehman, Jonathan Gordon, Shawn Jain, Kamal Ndousse, Cathy Yeh, and Kenneth~O Stanley.
\newblock Evolution through large models.
\newblock In \emph{Handbook of Evolutionary Machine Learning}, pp.\  331--366. Springer, 2023.

\bibitem[Li et~al.(2024)Li, Fox, and Goodman]{li2024automated}
Michael~Y Li, Emily~B Fox, and Noah~D Goodman.
\newblock Automated statistical model discovery with language models.
\newblock \emph{arXiv preprint arXiv:2402.17879}, 2024.

\bibitem[Li et~al.(2023)Li, Allal, Zi, Muennighoff, Kocetkov, Mou, Marone, Akiki, Li, Chim, et~al.]{starcoder}
Raymond Li, Loubna~Ben Allal, Yangtian Zi, Niklas Muennighoff, Denis Kocetkov, Chenghao Mou, Marc Marone, Christopher Akiki, Jia Li, Jenny Chim, et~al.
\newblock Starcoder: may the source be with you!
\newblock \emph{arXiv preprint arXiv:2305.06161}, 2023.

\bibitem[Liu et~al.(2023)Liu, Astorga, Seedat, and van~der Schaar]{llambo-llm-bayesian}
Tennison Liu, Nicol{\'a}s Astorga, Nabeel Seedat, and Mihaela van~der Schaar.
\newblock Large language models to enhance bayesian optimization.
\newblock In \emph{The Twelfth International Conference on Learning Representations}, 2023.

\bibitem[Ma et~al.(2024)Ma, Wang, Guo, Sun, Tenenbaum, Rus, Gan, and Matusik]{ma2024llm-bilevel}
Pingchuan Ma, Tsun-Hsuan Wang, Minghao Guo, Zhiqing Sun, Joshua~B Tenenbaum, Daniela Rus, Chuang Gan, and Wojciech Matusik.
\newblock Llm and simulation as bilevel optimizers: A new paradigm to advance physical scientific discovery.
\newblock \emph{arXiv preprint arXiv:2405.09783}, 2024.

\bibitem[Madaan et~al.(2024)Madaan, Tandon, Gupta, Hallinan, Gao, Wiegreffe, Alon, Dziri, Prabhumoye, Yang, et~al.]{self_refine-2024}
Aman Madaan, Niket Tandon, Prakhar Gupta, Skyler Hallinan, Luyu Gao, Sarah Wiegreffe, Uri Alon, Nouha Dziri, Shrimai Prabhumoye, Yiming Yang, et~al.
\newblock Self-refine: Iterative refinement with self-feedback.
\newblock \emph{Advances in Neural Information Processing Systems}, 36, 2024.

\bibitem[Majumder et~al.(2024)Majumder, Surana, Agarwal, Hazra, Sabharwal, and Clark]{data-driven-AI2-LM}
Bodhisattwa~Prasad Majumder, Harshit Surana, Dhruv Agarwal, Sanchaita Hazra, Ashish Sabharwal, and Peter Clark.
\newblock Data-driven discovery with large generative models.
\newblock \emph{arXiv preprint arXiv:2402.13610}, 2024.

\bibitem[Maza \& Tidor(1993)Maza and Tidor]{boltzman-selection}
Michael de~la Maza and Bruce Tidor.
\newblock An analysis of selection procedures with particular attention paid to proportional and boltzmann selection.
\newblock In \emph{Proceedings of the 5th International Conference on Genetic Algorithms}, pp.\  124–131, San Francisco, CA, USA, 1993. Morgan Kaufmann Publishers Inc.
\newblock ISBN 1558602992.

\bibitem[Meidani et~al.(2023)Meidani, Shojaee, Reddy, and Farimani]{SNIP-ICLR}
Kazem Meidani, Parshin Shojaee, Chandan~K Reddy, and Amir~Barati Farimani.
\newblock Snip: Bridging mathematical symbolic and numeric realms with unified pre-training.
\newblock In \emph{The Twelfth International Conference on Learning Representations}, 2023.

\bibitem[Meyerson et~al.(2023)Meyerson, Nelson, Bradley, Gaier, Moradi, Hoover, and Lehman]{LMX-Lehman-2023}
Elliot Meyerson, Mark~J Nelson, Herbie Bradley, Adam Gaier, Arash Moradi, Amy~K Hoover, and Joel Lehman.
\newblock Language model crossover: Variation through few-shot prompting.
\newblock \emph{arXiv preprint arXiv:2302.12170}, 2023.

\bibitem[Mundhenk et~al.(2021)Mundhenk, Landajuela, Glatt, Santiago, faissol, and Petersen]{mundhenk-seeding-GP-NeurIPS-2021}
Terrell~N. Mundhenk, Mikel Landajuela, Ruben Glatt, Claudio~P. Santiago, Daniel faissol, and Brenden~K. Petersen.
\newblock Symbolic regression via deep reinforcement learning enhanced genetic programming seeding.
\newblock In A.~Beygelzimer, Y.~Dauphin, P.~Liang, and J.~Wortman Vaughan (eds.), \emph{Advances in Neural Information Processing Systems}, 2021.

\bibitem[Petersen et~al.(2021)Petersen, Larma, Mundhenk, Santiago, Kim, and Kim]{DSR-Petersen-ICLR-2021}
Brenden~K Petersen, Mikel~Landajuela Larma, Terrell~N. Mundhenk, Claudio~Prata Santiago, Soo~Kyung Kim, and Joanne~Taery Kim.
\newblock Deep symbolic regression: Recovering mathematical expressions from data via risk-seeking policy gradients.
\newblock In \emph{International Conference on Learning Representations}, 2021.

\bibitem[Qi et~al.(2023)Qi, Zhang, Li, Tian, Zeng, Chen, and Zhou]{LLM-scientific-zero-shot-hypothesis}
Biqing Qi, Kaiyan Zhang, Haoxiang Li, Kai Tian, Sihang Zeng, Zhang-Ren Chen, and Bowen Zhou.
\newblock Large language models are zero shot hypothesis proposers.
\newblock \emph{arXiv preprint arXiv:2311.05965}, 2023.

\bibitem[Romera-Paredes et~al.(2024)Romera-Paredes, Barekatain, Novikov, Balog, Kumar, Dupont, Ruiz, Ellenberg, Wang, Fawzi, Kohli, and Fawzi]{FunSearch}
Bernardino Romera-Paredes, Mohammadamin Barekatain, Alexander Novikov, Matej Balog, M.~Pawan Kumar, Emilien Dupont, Francisco J.~R. Ruiz, Jordan~S. Ellenberg, Pengming Wang, Omar Fawzi, Pushmeet Kohli, and Alhussein Fawzi.
\newblock Mathematical discoveries from program search with large language models.
\newblock \emph{Nature}, 625\penalty0 (7995):\penalty0 468--475, Jan 2024.
\newblock ISSN 1476-4687.
\newblock \doi{10.1038/s41586-023-06924-6}.

\bibitem[Schmidt \& Lipson(2009)Schmidt and Lipson]{Schmidt-Lipson-2009}
Michael Schmidt and Hod Lipson.
\newblock Distilling free-form natural laws from experimental data.
\newblock \emph{Science Advance}, 324\penalty0 (5923):\penalty0 81--85, 2009.
\newblock ISSN 0036-8075.
\newblock \doi{10.1126/science.1165893}.

\bibitem[Shojaee et~al.(2023)Shojaee, Jain, Tipirneni, and Reddy]{rl_code1}
Parshin Shojaee, Aneesh Jain, Sindhu Tipirneni, and Chandan~K Reddy.
\newblock Execution-based code generation using deep reinforcement learning.
\newblock \emph{arXiv preprint arXiv:2301.13816}, 2023.

\bibitem[Shojaee et~al.(2024)Shojaee, Meidani, Barati~Farimani, and Reddy]{TPSR-NeurIPS}
Parshin Shojaee, Kazem Meidani, Amir Barati~Farimani, and Chandan Reddy.
\newblock Transformer-based planning for symbolic regression.
\newblock \emph{Advances in Neural Information Processing Systems}, 36, 2024.

\bibitem[Sun et~al.(2023)Sun, Liu, Wang, and Sun]{SPL-MCTS-ICLR-2023}
Fangzheng Sun, Yang Liu, Jian-Xun Wang, and Hao Sun.
\newblock Symbolic physics learner: Discovering governing equations via monte carlo tree search.
\newblock In \emph{The Eleventh International Conference on Learning Representations}, 2023.

\bibitem[Todorovski \& Dzeroski(1997)Todorovski and Dzeroski]{Todorovski-Declarative-SR}
Ljupco Todorovski and Saso Dzeroski.
\newblock Declarative bias in equation discovery.
\newblock In \emph{Proceedings of the Fourteenth International Conference on Machine Learning}, ICML '97, pp.\  376–384, San Francisco, CA, USA, 1997. Morgan Kaufmann Publishers Inc.
\newblock ISBN 1558604863.

\bibitem[Todorovski \& D{\v{z}}eroski(2007)Todorovski and D{\v{z}}eroski]{Todorovski-integrate-SR}
Ljup{\v{c}}o Todorovski and Sa{\v{s}}o D{\v{z}}eroski.
\newblock \emph{Integrating Domain Knowledge in Equation Discovery}, pp.\  69--97.
\newblock Springer Berlin Heidelberg, Berlin, Heidelberg, 2007.
\newblock ISBN 978-3-540-73920-3.
\newblock \doi{10.1007/978-3-540-73920-3\_4}.

\bibitem[Udrescu \& Tegmark(2020)Udrescu and Tegmark]{AI-Feynman-Science-2020}
Silviu-Marian Udrescu and Max Tegmark.
\newblock Ai feynman: A physics-inspired method for symbolic regression.
\newblock \emph{Science Advances}, 6\penalty0 (16):\penalty0 eaay2631, 2020.
\newblock \doi{10.1126/sciadv.aay2631}.

\bibitem[Virgolin \& Pissis(2022)Virgolin and Pissis]{NP-hard-Symbolic-2022}
Marco Virgolin and Solon~P Pissis.
\newblock Symbolic regression is {NP}-hard.
\newblock \emph{Transactions on Machine Learning Research}, 2022.
\newblock ISSN 2835-8856.

\bibitem[Wang et~al.(2023{\natexlab{a}})Wang, Fu, Du, Gao, Huang, Liu, Chandak, Liu, Van~Katwyk, Deac, Anandkumar, Bergen, Gomes, Ho, Kohli, Lasenby, Leskovec, Liu, Manrai, Marks, Ramsundar, Song, Sun, Tang, Veli{\v{c}}kovi{\'{c}}, Welling, Zhang, Coley, Bengio, and Zitnik]{scientific-age-AI}
Hanchen Wang, Tianfan Fu, Yuanqi Du, Wenhao Gao, Kexin Huang, Ziming Liu, Payal Chandak, Shengchao Liu, Peter Van~Katwyk, Andreea Deac, Anima Anandkumar, Karianne Bergen, Carla~P. Gomes, Shirley Ho, Pushmeet Kohli, Joan Lasenby, Jure Leskovec, Tie-Yan Liu, Arjun Manrai, Debora Marks, Bharath Ramsundar, Le~Song, Jimeng Sun, Jian Tang, Petar Veli{\v{c}}kovi{\'{c}}, Max Welling, Linfeng Zhang, Connor~W. Coley, Yoshua Bengio, and Marinka Zitnik.
\newblock Scientific discovery in the age of artificial intelligence.
\newblock \emph{Nature}, 620\penalty0 (7972):\penalty0 47--60, Aug 2023{\natexlab{a}}.
\newblock ISSN 1476-4687.
\newblock \doi{10.1038/s41586-023-06221-2}.

\bibitem[Wang et~al.(2023{\natexlab{b}})Wang, Zelikman, Poesia, Pu, Haber, and Goodman]{wang2023hypothesis}
Ruocheng Wang, Eric Zelikman, Gabriel Poesia, Yewen Pu, Nick Haber, and Noah~D Goodman.
\newblock Hypothesis search: Inductive reasoning with language models.
\newblock \emph{arXiv preprint arXiv:2309.05660}, 2023{\natexlab{b}}.

\bibitem[Wu et~al.(2024)Wu, Wu, Wu, Feng, and Tan]{LLM-Evo-Survey}
Xingyu Wu, Sheng-hao Wu, Jibin Wu, Liang Feng, and Kay~Chen Tan.
\newblock Evolutionary computation in the era of large language model: Survey and roadmap.
\newblock \emph{arXiv preprint arXiv:2401.10034}, 2024.

\bibitem[Wu et~al.(2023)Wu, Qiu, Ross, Aky{\"u}rek, Chen, Wang, Kim, Andreas, and Kim]{wu2023reasoning}
Zhaofeng Wu, Linlu Qiu, Alexis Ross, Ekin Aky{\"u}rek, Boyuan Chen, Bailin Wang, Najoung Kim, Jacob Andreas, and Yoon Kim.
\newblock Reasoning or reciting? exploring the capabilities and limitations of language models through counterfactual tasks.
\newblock \emph{arXiv preprint arXiv:2307.02477}, 2023.

\bibitem[Yang et~al.(2023{\natexlab{a}})Yang, Wang, Lu, Liu, Le, Zhou, and Chen]{LLM-Optimizer}
Chengrun Yang, Xuezhi Wang, Yifeng Lu, Hanxiao Liu, Quoc~V Le, Denny Zhou, and Xinyun Chen.
\newblock Large language models as optimizers.
\newblock \emph{arXiv preprint arXiv:2309.03409}, 2023{\natexlab{a}}.

\bibitem[Yang et~al.(2023{\natexlab{b}})Yang, Swope, Gu, Chalamala, Song, Yu, Godil, Prenger, and Anandkumar]{LeanDojo-theorem}
Kaiyu Yang, Aidan Swope, Alex Gu, Rahul Chalamala, Peiyang Song, Shixing Yu, Saad Godil, Ryan Prenger, and Anima Anandkumar.
\newblock {LeanDojo}: Theorem proving with retrieval-augmented language models.
\newblock In \emph{Neural Information Processing Systems (NeurIPS)}, 2023{\natexlab{b}}.

\bibitem[Zheng et~al.(2023{\natexlab{a}})Zheng, Su, You, Wang, Qian, Xu, and Albanie]{gpt4-NAS-llm}
Mingkai Zheng, Xiu Su, Shan You, Fei Wang, Chen Qian, Chang Xu, and Samuel Albanie.
\newblock Can gpt-4 perform neural architecture search?
\newblock \emph{arXiv preprint arXiv:2304.10970}, 2023{\natexlab{a}}.

\bibitem[Zheng et~al.(2023{\natexlab{b}})Zheng, Koh, Ju, Nguyen, May, Webb, and Pan]{LLM-scientific-synthesis}
Yizhen Zheng, Huan~Yee Koh, Jiaxin Ju, Anh~TN Nguyen, Lauren~T May, Geoffrey~I Webb, and Shirui Pan.
\newblock Large language models for scientific synthesis, inference and explanation.
\newblock \emph{arXiv preprint arXiv:2310.07984}, 2023{\natexlab{b}}.

\end{thebibliography}
\bibliographystyle{iclr2025_conference}

\newpage 

\appendix 
\section*{Appendix}

\section{Baseline Implementation Details}
\label{sec:app-imp}
\vspace{-0.8em}
\subsection{Models}
\vspace{-0.8em}
We compare \modelname against several state-of-the-art Symbolic Regression (SR) baselines, encompassing a diverse range of methodologies from traditional evolutionary approaches to modern deep learning-based techniques. 
The baselines include:

\vspace{-0.8em}
\paragraph{GPlearn}
GPlearn is a pioneering and standard genetic programming (GP) SR approach. We use the open-source \texttt{gplearn}\footnote{\url{https://gplearn.readthedocs.io/en/stable/}} package with the following parameters: Population size: 500, Tournament size: 20, Maximum generations: 2 million. Most of the hyperparameters are set from default setting. 

\vspace{-0.8em}
\paragraph{PySR}
PySR \citep{pysr-2023} is an advanced SR method that employs asynchronous multi-island GP-based evolutions. We implement PySR using the open-source \texttt{pysr}\footnote{\url{https://github.com/MilesCranmer/PySR}} package with the following settings: Number of populations: 15, Population size: 33, Maximum iterations: 2 million. Except for the number of iterations, other parameters are the same as the default setting in PySR. This configuration leverages the power of parallel evolution over a long time allowing for a diverse and robust search of the equation space.

\vspace{-0.8em}
\paragraph{Deep Symbolic Regression (DSR)}
DSR \citep{DSR-Petersen-ICLR-2021} employs an RNN-based reinforcement learning search over symbolic expressions. We implement DSR using the open-source \texttt{deep-symbolic-optimization (DSO)}\footnote{\url{https://github.com/dso-org/deep-symbolic-optimization}} package with standard default parameters: Learning rate: 0.0005, Batch size: 512, and Maximum iterations: 2 million. This approach allows for a guided search through the space of symbolic expressions, leveraging the power of deep learning to inform the exploration process.

\vspace{-0.8em}
\paragraph{Unified Deep Symbolic Regression (uDSR)}
uDSR \citep{UDSR-NeurIPS-2022} extends DSR by incorporating additional linear token and GP search at the decoding stage. We also implement uDSR using the \texttt{DSO} package with the same default parameters as DSR. This unified approach aims to combine the strengths of deep learning and traditional GP methods.

\vspace{-0.8em}
\paragraph{Neural Symbolic Regression that Scales (NeSymReS)}
NeSymReS \citep{Biggio-NeSymReS-ICML-2021} is the pioneering pre-trained Transformer SR model for expression skeleton generation. We implement it using the \texttt{NeuralSymbolicRegressionThatScales}\footnote{\url{https://github.com/SymposiumOrganization/NeuralSymbolicRegressionThatScales}} repository with the following default parameters: Number of datapoints passed to Transformer: 500, and Expression sampling size: 32. It is important to note that this model is limited to pre-training with $\leq 3$ variables. Consequently, we only apply the model to datasets with $d_{max}=3$, excluding the Bacterial Growth problem (which has 4 variables) for evaluation of this model.

\vspace{-0.8em}
\paragraph{End-to-End Symbolic Regression (E2E)}
E2E \citep{Kamienny-E2E-symbolic-NIPS-2022} is a more recent end-to-end pre-trained Transformer SR approach. We implement it using the \texttt{symbolicregression}\footnote{\url{https://github.com/facebookresearch/symbolicregression}} Facebook repository with the following default parameters: Number of datapoints passed to Transformer: 200, and Expression sampling size: 10. This model is also pre-trained for problems with $\leq 10$ variables.

\vspace{-0.8em}
\subsection{Data Preprocessing and Model Execution}
\vspace{-0.5em}
For the pre-trained Transformer SR models (NeSymReS and E2E), data normalization is crucial. We apply standard normalization to the input data before feeding it to these Transformer models to ensure optimal performance.
For the search-based methods, we allow all baselines (GPlearn, PySR, DSR, and uDSR) to run for over 2 million iterations until convergence to their best performance. 
In the experiments, each baseline undergoes 5 replications. The best results obtained were then documented and reported. This extensive evaluation process, with a large number of iterations and search evaluations, ensures a robust assessment of each model's capability to converge towards optimal solutions and effectively explore the vast equation space of each problem.

\section{Details of LLM-SR Method and Implementation}
\label{sec:app-method}


\paragraph{Hypothesis Generation and Data-driven Evaluation}

\begin{figure}[t]
\centering
\includegraphics[width=0.70\linewidth]{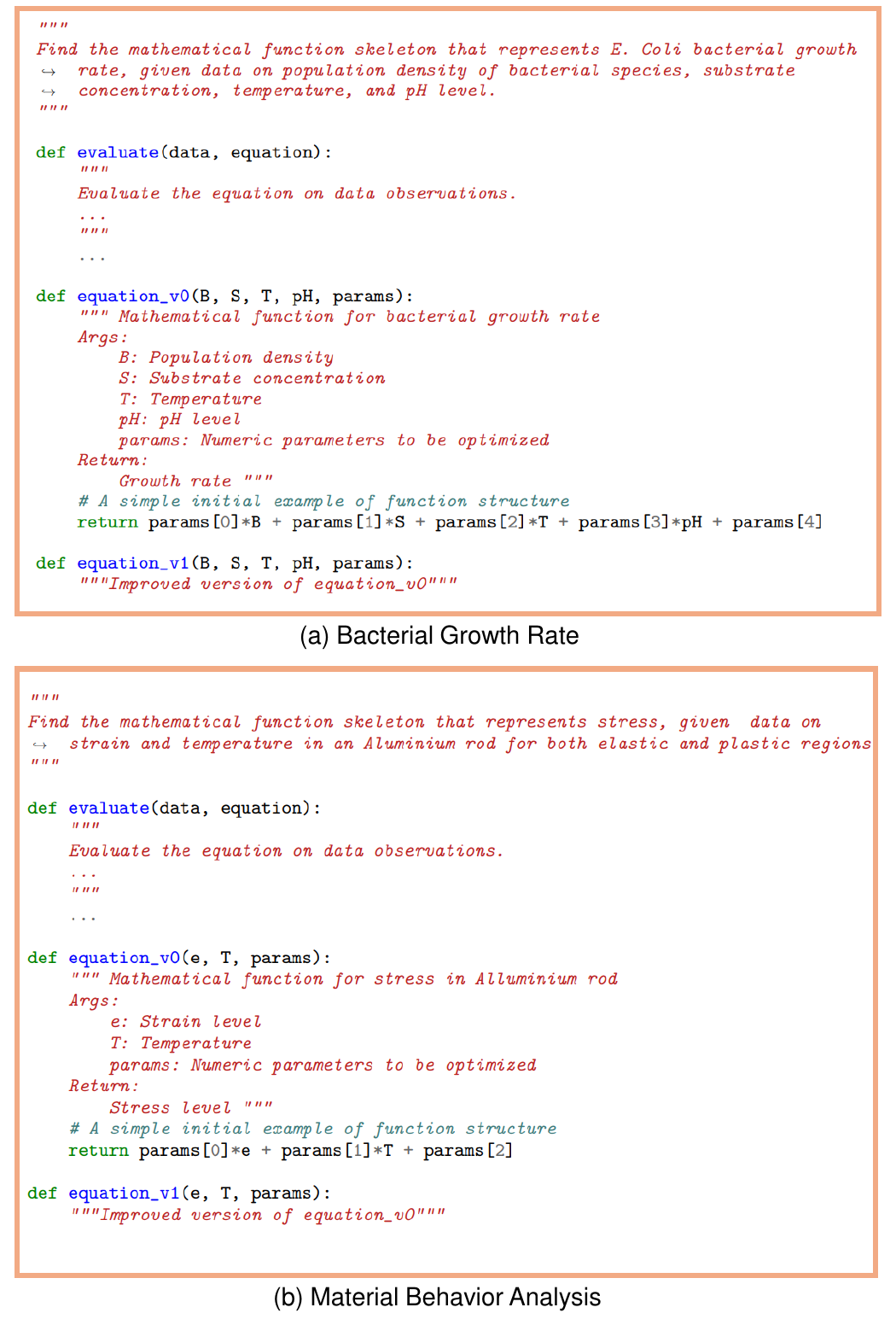}
\caption{\small Example of input prompts program body for \textbf{(a)} E. Coli Growth and \textbf{(b)} Stress-Strain problems, with problem specification, and the initial equation program example (set as simple linear equation skeleton). For better readability, the details of evaluation function are not included in this figure. Check Fig.~\ref{fig:spec-bacteria} for details.
\vspace{-0.5em}
}
\vspace{-0.5em}
\label{fig:spec-prompts}
\end{figure}

\begin{figure}[t]
\centering
\includegraphics[width=0.75\linewidth]{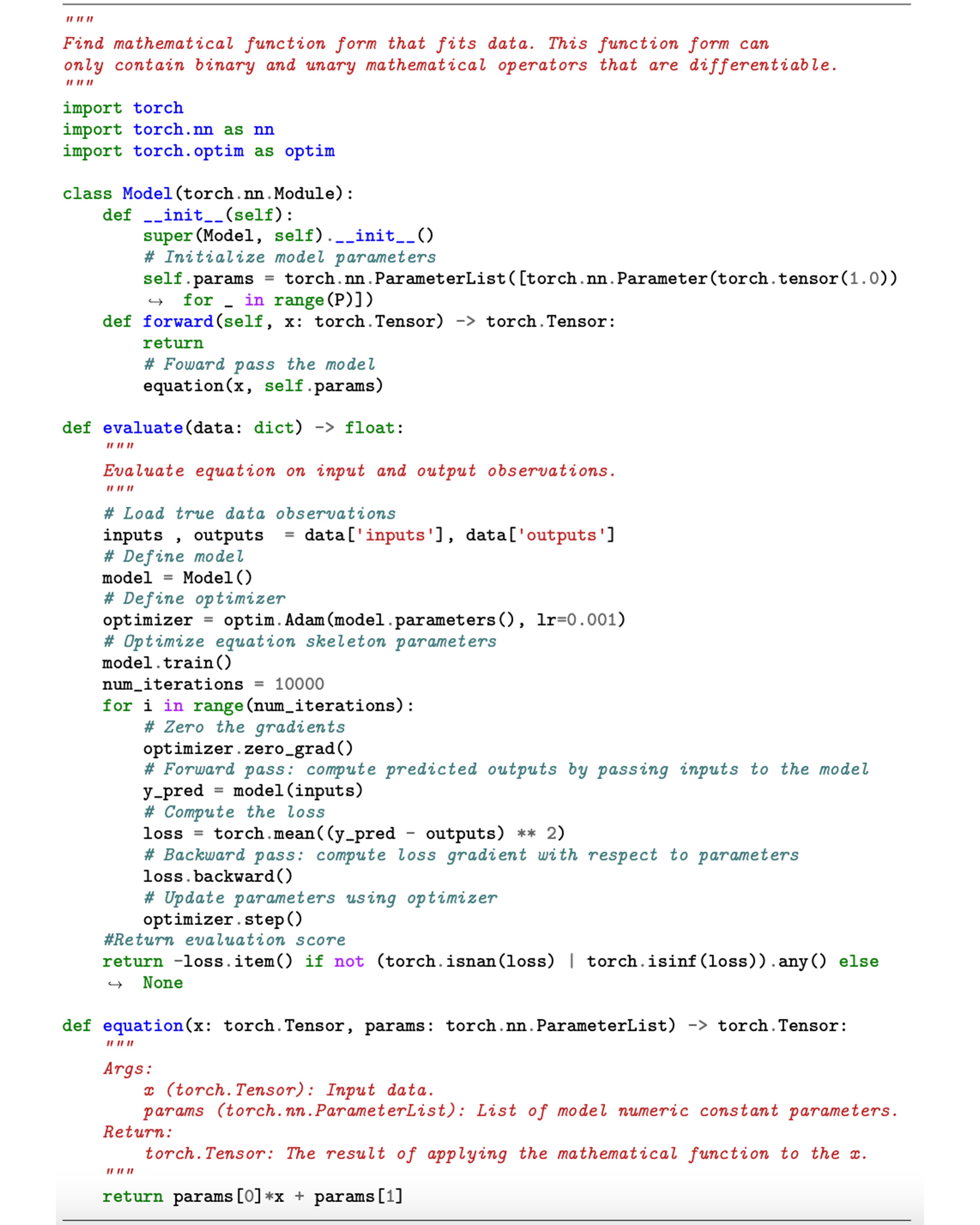}
\caption{An example of prompt structure, containing problem specification, evaluation and optimization function, and equation program with \texttt{pytorch} tensor operations.}
\label{fig:spec-torch}
\end{figure}

Fig.~\ref{fig:spec-bacteria} provided an example of specification for Nonlinear Oscillator problem. Here, Fig.~\ref{fig:spec-prompts} showcases illustrative examples of prompts and specifications tailored for the Bacterial Growth and Stress-Strain problems. These prompts contain descriptions of the problem and relevant variables, expressed in natural language. By providing this context, the language model can leverage its existing domain knowledge about the physical meaning and relations of variables to generate scientifically plausible hypotheses for new equation programs. Fig.~\ref{fig:spec-torch} also shows a more detailed example of prompt and specification for \modelname that prompts the model to generate differentiable equation programs in \texttt{PyTorch} using tensor operations. 
The prompt suggests using differentiable operators and replacing non-differentiable components (e.g., if-else conditions) with smooth differentiable approximations.

Our experiments employ either Mixtral-8x7B (using 4 NVIDIA RTX 8000 GPUs with 48GB memory each) or GPT-3.5-turbo (via OpenAI API) as the language model backbone. During each prompting step, the language model generates $b=4$ distinct equation program skeletons using a generation temperature of $\tau=0.8$. This temperature setting is chosen based on preliminary experiments to balance creativity (exploration) and adherence to the problem constraints and reliance on the prior knowledge (exploitation). To control the length and the complexity of the generated equations and prevent overparameterization, we set the maximum number of parameters (length of \texttt{params} vector) as $10$ in all experiments. The generated equation skeleton programs are then evaluated to gather feedback. In this framework, we deploy $e=4$ evaluators to operate concurrently. This parallelization allows for rapid and efficient assessment of the generated programs per prompt. Evaluation is constrained by time and memory limits set at $T = 30$ seconds and $M = 2$GB, respectively. Equation programs that exceed these limits are disqualified and considered as discarded hypotheses by returning \texttt{None} scores. This constraint ensures timely progress and resource efficiency in the search process.

\paragraph{Experience Buffer Management}
The system stores equation hypotheses and their corresponding data-driven scores in an experience buffer. It uses an islands model with multiple populations ($m=10$ islands) to maintain diversity. Each island is initialized with a simple equation, which can be customized for domain-specific problems. At each iteration, new hypotheses and their fitness scores are added to their originating island if they improve upon the island's best score.
To maintain the quality and diversity of the experience buffer, we follow \citep{FunSearch} and periodically reset the worst-performing islands. Every $T_{reset}$ iterations (every $4$ hrs), we identify the $m/2$ islands whose best equation programs have the lowest fitness scores.
All the equation programs in these islands are discarded, and each island is reinitialized with a single high-performing equation program, obtained by randomly selecting one of the surviving $m/2$ islands and copying its highest-scoring equation program (favoring older programs in case of ties). This reset mechanism allows the framework to discard stagnant or unproductive regions of the equation program space and focus on more promising areas.
Within each island, we further cluster the equation programs based on their signature, which is defined as the equation program score. Equation programs with identical signatures are grouped together, forming clusters within each island. This clustering approach helps preserve diversity by ensuring that equation programs with different performance characteristics are maintained in each population.

\paragraph{Experience Sampling}
To construct informative prompts for the LLM, we sample equation programs from the experience buffer and update the prompt to include new experience demonstration in-context examples. Similar to \citep{FunSearch}, here we use a two-stage sampling process. First, we randomly select an island from the $m$ available islands.  Then, within the selected island, we sample $k$ equation programs (typically, $k=2$) to be included as in-context examples in the prompt. 
When sampling equation programs within an island, we employ a two-step approach. First, we sample a cluster based on its evaluation score, favoring clusters with higher scores (i.e., higher-quality equation programs). Let $s_i$ denote the score of the $i$-th cluster, defined as an aggregation (e.g., mean) of all the scores in the signature that characterizes that cluster. The probability $P_i$ of choosing cluster $i$ is given by:
$$P_i = \frac{\exp\left( \frac{s_i }{\tau_{c}}\right)}{\sum_{i'} \exp\left( 
\frac{s_{i'}}{\tau_{c}}\right)}
, \quad \tau_{c} = T_0 \left(1 - \frac{u \bmod N}{N}\right),$$
where $\tau_{c}$ is the temperature parameter, $u$ is the current number of equation programs in the island, and $T_0=0.1$ and $N=10,000$ are hyperparameters. This selection approach is known as the Boltzmann selection procedure \citep{boltzman-selection}. Once a cluster is selected, we sample an equation program within that cluster, favoring shorter programs. Let $l_i$ denote the negative length of the $i$-th program within the chosen cluster (measured as the number of characters), and let $\tilde{l}_i = \frac{l_i - \min_{i'}{l_{i'}}}{\max_{i'}{l_{i'}} + 10^{-6}}$. We set the probability of selecting each equation program proportional to $\exp\left(-\tilde{l}_i / \tau_{p}\right)$, where $\tau_{p}=1$ is a temperature hyperparameter. The sampled programs are then included in the prompt as in-context experience demonstration, providing the LLM with relevant and diverse examples to guide the generation of new equation programs.
By maintaining a diverse and high-quality population in the experience buffer and employing a strategic sampling approach, the experience management enables the \modelname framework to effectively explore the space of equation programs and iteratively refine its search based on the most promising candidates.


\vspace{-0.8em}
\section{Limitation of Feynman Benchmark Problems}
\label{sec:app-feyn}
\vspace{-0.8em}

The Feynman benchmark \citep{AI-Feynman-Science-2020}, consisting of $120$ fundamental physics problems from the Feynman Lectures on Physics, is widely used to evaluate symbolic regression techniques in scientific equation discovery. However, our investigation indicates that LLMs have likely memorized many of these well-known physics equations. This memorization poses a challenge when using the Feynman benchmark to assess LLM-based equation discovery methods, as it may not accurately reflect the models' true discovery capabilities. This section elaborates on these limitations and provides evidence supporting the necessity of our newly designed benchmark problems.


\begin{wrapfigure}[11]{r}{0.45\textwidth}
\centering
\vspace{-1.5em}
\includegraphics[width=\linewidth]{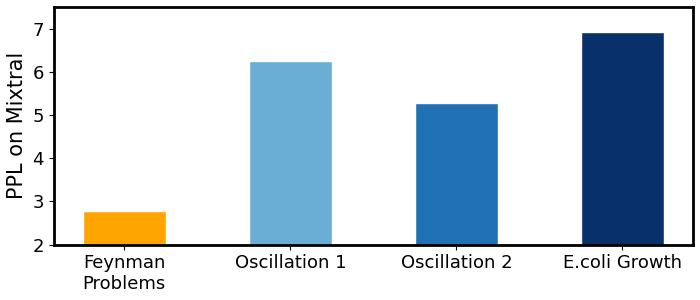}
\caption{\small Perplexity (\texttt{Mixtral}) comparison of Feynman benchmark and our new designed benchmark problems
\vspace{-0.5em}
}
\label{fig:feyn_perplexity}
\vspace{-0.5em}
\end{wrapfigure}
\begin{figure}[t]
\centering
\includegraphics[width=\linewidth]{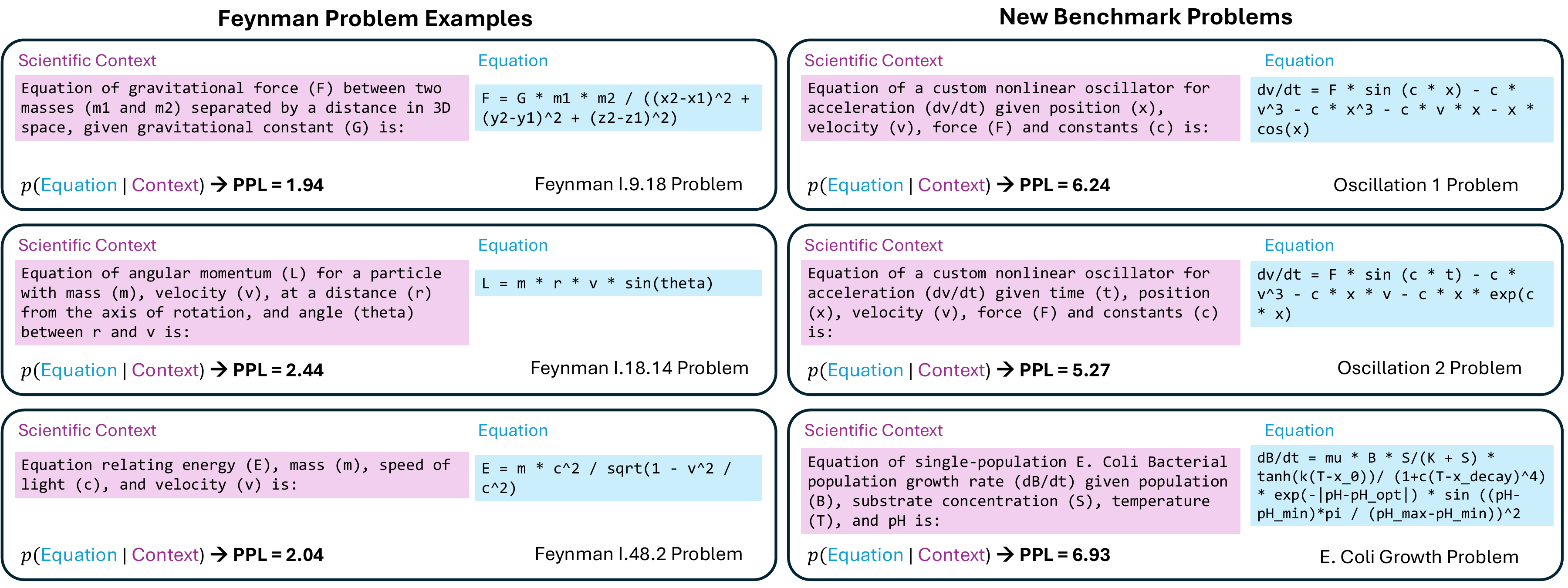}
\caption{\small Examples of input context and output equations for Feynman equations and our new benchmark problems in Perplexity experiments with Mixtral LLM backbone.
\vspace{-0.5em}
}
\label{fig:qual-perplexity}
\vspace{-0.5em}
\end{figure}

\paragraph{Perplexity Analysis}
To quantify the potential memorization of Feynman problems by LLMs, we first conducted a comparative perplexity analysis. Fig.~\ref{fig:feyn_perplexity} illustrates the median perplexity of Feynman problems against our new benchmarks (Oscillation 1, Oscillation 2, and E. Coli Growth) using the Mixtral-8x-7B model as the LLM backbone. 
Perplexity, in this context, is calculated only for the generation of equations given scientific context: $p(\text{Equation}|\text{Context})$. Examples of input prompts and outputs used for perplexity computation across different benchmarks are provided in Fig.~\ref{fig:qual-perplexity}. Mathematically, perplexity is defined as:
$\text{PPL} = \exp\left(-\frac{1}{N}\sum_{i=1}^N \log p(x_i|x_{<i})\right)$, where $N$ is the number of tokens in the generated equation, and $p(x_i|x_{<i})$ is the probability of token $x_i$ given the preceding tokens, derived from the logits of the Mixtral model. The significantly lower perplexity observed for Feynman problems indicates a higher certainty in the LLM's predictions for these equations, and a higher chance of LLM recitation rather than reasoning and discovery. This suggests a high likelihood of memorization of well-known Feynman equations by the LLM, potentially due to their prevalence in scientific training data.
It is worth noting that we have excluded the Stress-Strain problem from this analysis due to its experimental nature and lack of a predetermined theoretical model structure, precluding the calculation of perplexity.

\begin{wrapfigure}[21]{r}{0.5\textwidth}
\centering
\vspace{-1.0em}
\includegraphics[width=\linewidth]{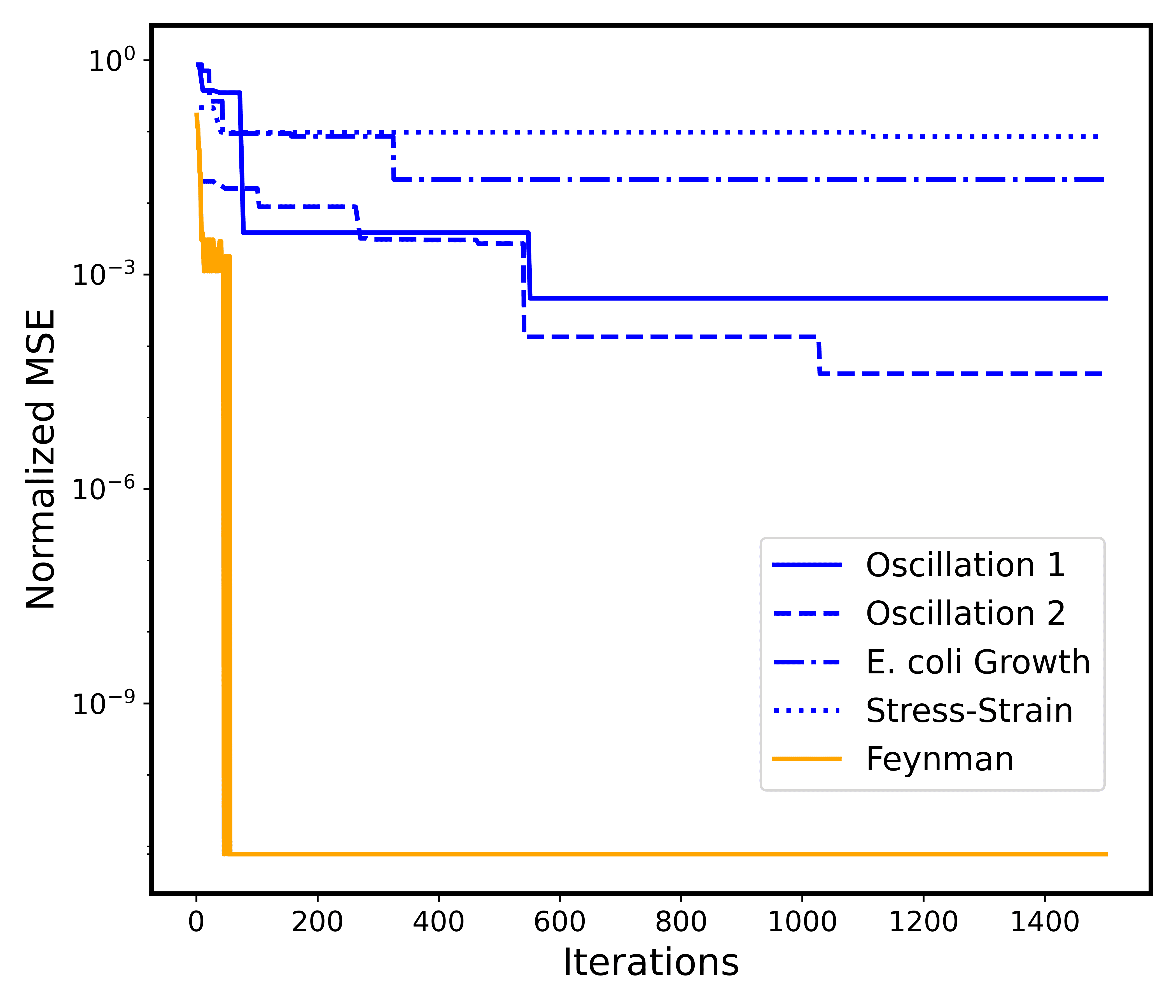}
\caption{\small Trajectory of Normalized MSE score over iterations for \modelname (\texttt{GPT-3.5}) on Feynman benchmark problems versus new benchmarks
\vspace{-0.5em}
}
\label{fig:feyn-all}
\vspace{-0.5em}
\end{wrapfigure}

\begin{figure}[t]
\centering
\includegraphics[width=.75\linewidth]{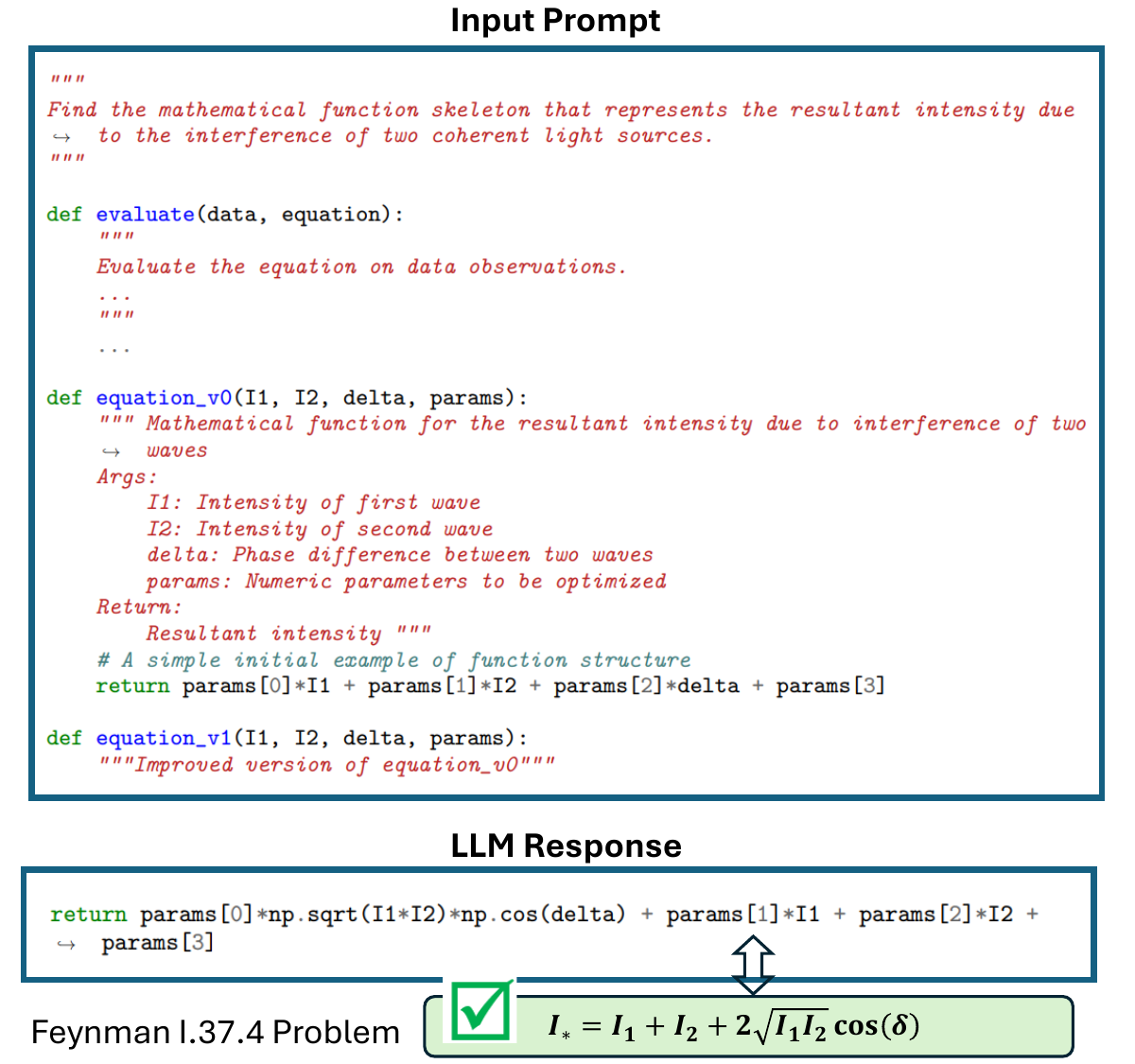}
\caption{\small An example of LLM response to Feynman I.37.4 problem, demonstrating LLM recitation without iterative search. For better readability, the details of evaluation function are not included in this figure. Check Fig.~\ref{fig:spec-bacteria} for details.
}
\vspace{-0.5em}
\label{fig:qual-feynI374}
\vspace{-0.5em}
\end{figure}

\begin{figure}[t]
\centering
\includegraphics[width=0.75\linewidth]{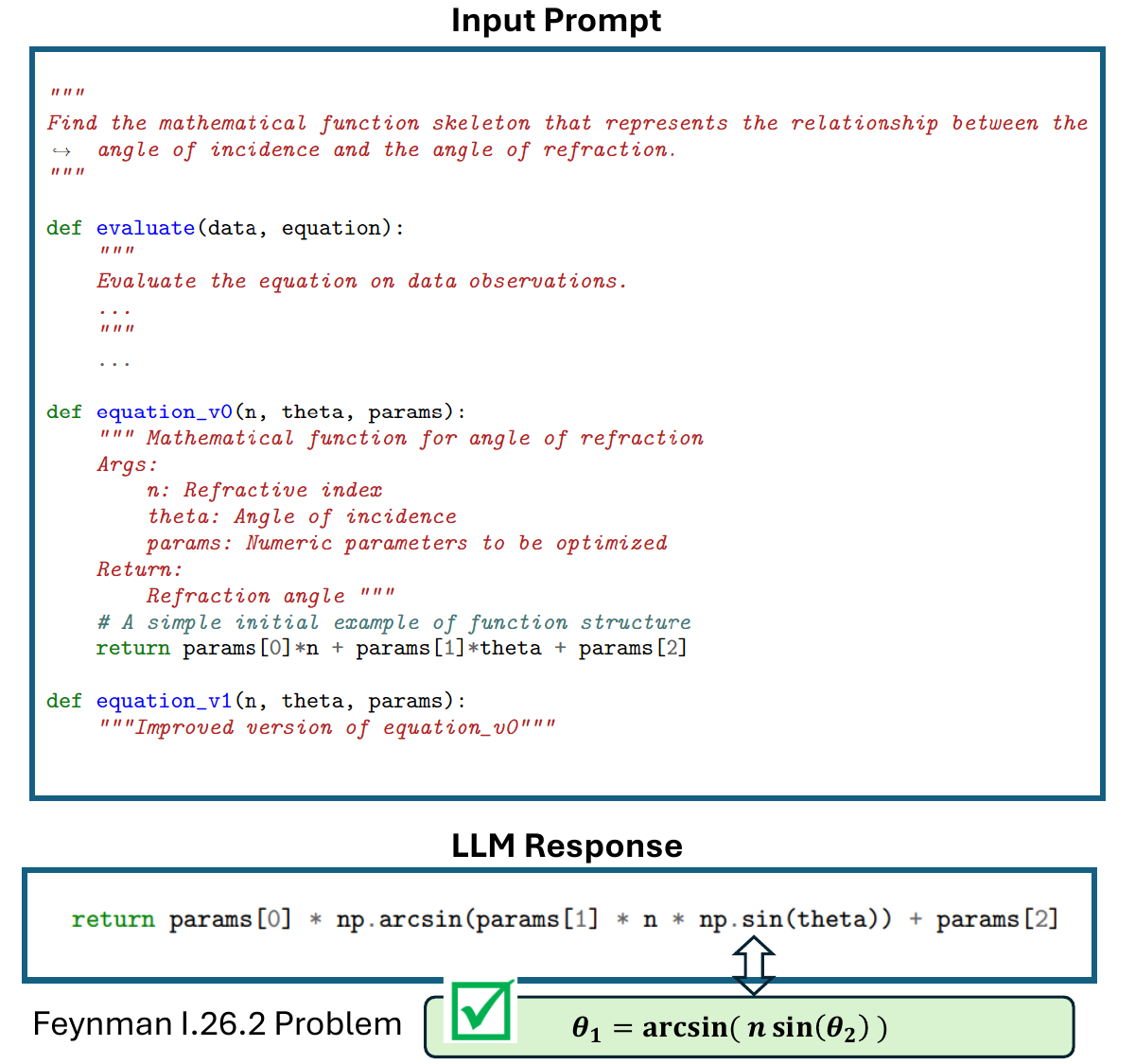}
\caption{\small An example of LLM response to Feynman I.26.2 problem, demonstrating LLM recitation without iterative search. For better readability, the details of evaluation function are not included in this figure. Check Fig.~\ref{fig:spec-bacteria} for details.
}
\vspace{-0.5em}
\label{fig:qual-feynI262}
\vspace{-0.5em}
\end{figure}

\paragraph{Discovery Error Curve Analysis}
To further validate the need for new benchmarks, we compared equation discovery error curves between Feynman problems and our new benchmark problems. 
Fig.~\ref{fig:feyn-all} presents the performance of \modelname with a GPT-3.5 backbone across various problems, showing the best score trajectory of Normalized Mean Squared Error (NMSE) against the number of iterations. For Feynman benchmark problems, \modelname achieves low NMSE scores within very few iterations, often in a single pass. This rapid convergence further supports the hypothesis that LLMs have likely memorized these fundamental physics equations due to their ubiquity in training data.
Qualitative examples in Figs.~\ref{fig:qual-feynI374} and \ref{fig:qual-feynI262} also provide additional evidence. The LLM's one-pass responses to several Feynman problems not only demonstrate functional accuracy but also often recite the exact form of the corresponding physics expressions, suggesting direct recall rather than a discovery process.
In contrast, our newly designed benchmark problems present novel challenges requiring reasoning and exploration.


\section{Additional Details on New Benchmark Problems}
\label{sec:app-data}

The datasets used in this study include both publicly available and newly generated data. The material stress behavior analysis dataset (stress-strain) is publicly available under the CC BY 4.0 license and can be accessed at \href{https://data.mendeley.com/datasets/rd6jm9tyb6/1}{https://data.mendeley.com/datasets/rd6jm9tyb6/1}. The remaining datasets (Oscillation 1, Oscillation 2, and E. coli Growth) were generated for this work and are released under the MIT License as part of the LLM-SR GitHub repository: \href{https://github.com/deep-symbolic-mathematics/LLM-SR}{https://github.com/deep-symbolic-mathematics/LLM-SR}

\subsection{Nonlinear Oscillator Equations}
\label{sec:app-oscillator}

In this work, we introduce two novel nonlinear oscillator systems as part of our effort to create more challenging and robust benchmarks for LLM-based equation discovery methods. These custom-designed oscillators extend beyond commonly studied systems like Van der Pol, Rayleigh, or Duffing oscillators, presenting unique challenges that test the reasoning and discovery capabilities of LLM-based approaches. 

The general form of nonlinear damped oscillator equations is typically expressed as: $\ddot{x} + f(t, x, \dot{x}) = 0$, 
where $t$ represents time, $x$ represents position, and $f(t,x, \dot{x})$ represents nonlinear forces. Our custom designs expand upon this framework, incorporating a rich combination of nonlinear terms to create systems that are challenging yet solvable.
We simulate two nonlinear oscillators using the \texttt{solve\_ivp} function from the \texttt{scipy} library to generate data.

\begin{table}[htp]
\centering
\resizebox{0.6\linewidth}{!}{
\begin{tabular}{lcccccccc}
\toprule
Dataset      & Time range & initial values   & F   & $\alpha$ & $\beta$ & $\delta$ & $\gamma$ & $\omega$ \\
\midrule
Oscillator 1 & (0, 50)    & \{x=0.5, v=0.5\} & 0.8 & 0.5      & 0.2     & \_       & 0.5      & 1.0      \\
Oscillator 2 & (0, 50)    & \{x=0.5, v=0.5\} & 0.3 & 0.5      & 1.0     & 5.0      & 0.5      & 1.0       \\
\bottomrule
\end{tabular}
}
\caption{Parameter values for Oscillator datasets.}
\label{table_oscillator_params}
\end{table}

The parameters and initial values for these simulations are provided in Table~\ref{table_oscillator_params}. The governing equations for our oscillator systems are as follows:

\textbf{Oscillator 1:}
$$\dot{v} = F \sin(\omega x) -\alpha v^3 - \beta x^3 -\gamma x \cdot v - x \cos(x)$$

\textbf{Oscillator 2:}
$$\dot{v} = F \sin(\omega t) - \alpha v^3 - \beta x \cdot v - \delta x \cdot \exp(\gamma x)$$

where $v = \dot{x}$ represents velocity, and $F, \omega, \alpha, \beta, \gamma, \delta$ are constants specific to each oscillator system. These equations are carefully deisgned to incorporate a diverse set of nonlinear structures, including trigonometric, polynomial, and exponential terms. This design choice serves multiple purposes:

\begin{itemize}[leftmargin=*]
\item \textbf{Challenging Complexity}: The combination of various nonlinear terms creates a rich dynamical system that is more complex than common oscillator systems, making the equation discovery task non-trivial.
\item \textbf{Realistic Physics}: While complex, these equations are still solvable and still represent physically plausible systems, incorporating recognizable elements such as nonlinear damping and position-dependent restoring forces.
\item \textbf{Novelty}: By deviating from well-known oscillator forms, we reduce the likelihood of LLMs simply reciting memorized equations, thus testing their reasoning and discovery capabilities in the context of data-driven scientific equation discovery.
\end{itemize}

\begin{wrapfigure}[15]{r}{0.45\textwidth}
\centering
\includegraphics[width=\linewidth]{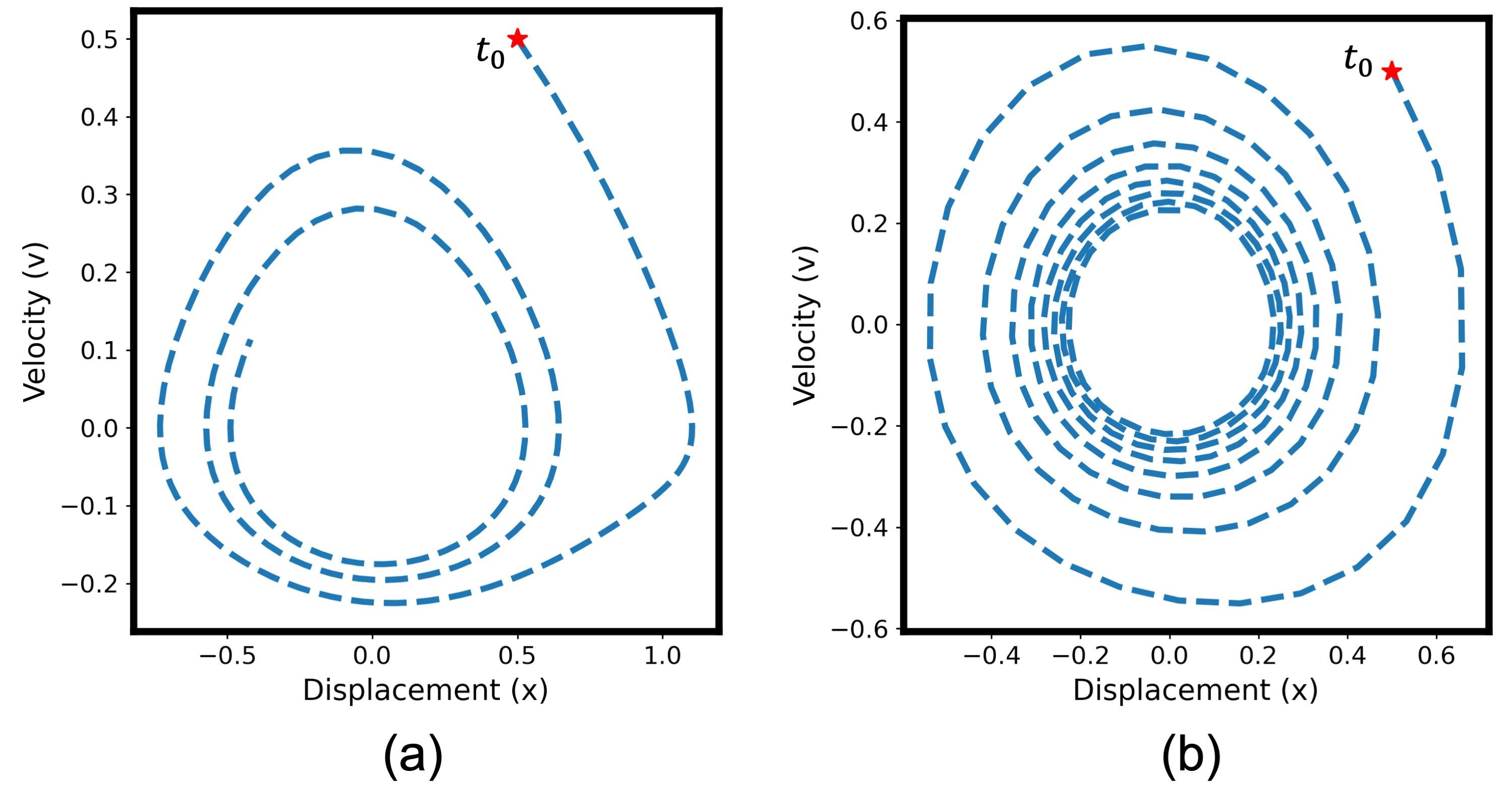}
\caption{\small Phase diagrams of trajectories corresponding to custom oscillators: \textbf{(a)}~Oscillator~1 and \textbf{(b)}~Oscillator~2}
\label{fig:oscillation_phase}
\vspace{-1em}
\end{wrapfigure}

Fig.~\ref{fig:oscillation_phase} illustrates the phase plane diagrams of these nonlinear damped oscillators, visually demonstrating the complex dynamics arising from the interplay of nonlinear driving forces, restoring forces, and damping forces. 
These diagrams highlight the rich behavior that makes these systems challenging for equation identification tasks.
To effectively evaluate the generalization capability of predicted equations, we employ a strategic data partitioning scheme. The simulation data is divided into three sets based on the trajectory time: (1)~Training set, (2)~In-domain validation set, and (3)~Out-of-domain validation set.
Specifically, we utilize the time interval $T=[0,20)$ to evaluate the out-of-domain generalization of the discovered equations. 

\subsection{E. coli Growth Rate Equations}
\label{sec:app-bacterial}
In the domain of microbiology, understanding and modeling the growth dynamics of Escherichia coli (E. coli) is of paramount importance due to its wide-ranging applications in biotechnology, food safety, and fundamental biological research. To advance the LLM-based equation discovery approaches in this field, we have developed a novel benchmark problem centered around E. coli growth rate modeling.
The growth rate of bacterial populations, including E. coli, is typically modeled by a differential equation that incorporates multiple environmental factors. This mathematical model commonly takes a multiplicative form:

$$\frac{dB}{dt} = f(B,S,T,\text{pH}) = f_B(B) \cdot f_S(S) \cdot f_T(T) \cdot f_{\text{pH}}(\text{pH}),$$

where $B$ represents bacterial population density, $S$ is substrate concentration, $T$ is temperature, and $\text{pH}$ represents the acidity or alkalinity of the growth medium.
To create a benchmark that is grounded in biological prior knowledge yet is challenging, we have extended this framework with a custom differential equation:

{\small $$
\frac{dB}{dt} = \mu_{max} B \left(\frac{S}{K_s + S}\right) \frac{\tanh\left(k(T-x_0)\right)}{1+c(T-x_{decay})^4} \exp\left(-\left|\text{pH}-\text{pH}_{opt}\right|\right) \sin\left(\frac{(\text{pH}-\text{pH}_{min})\pi}{\text{pH}_{max}-\text{pH}_{min}}\right)^2
$$}

This equation incorporates several key components, each designed to test different aspects of equation discovery systems:

\begin{itemize}[leftmargin=*]
\item \textbf{Population Density} ($f_B$): We maintain a linear relationship with $B$, reflecting a simple single-population scenario. This choice allows the focus to remain on the more complex environmental dependencies.
\item \textbf{Substrate Concentration} ($f_S$): We employ the well-established Monod equation, $\left(\frac{S}{K_s + S}\right)$, which has been a cornerstone of bacterial growth modeling since its introduction by Jacques Monod in 1949. This inclusion serves as prior knowledge, allowing us to evaluate how well discovery methods can identify known relationships within a more complex overall structure.
\item \textbf{Temperature Dependency} ($f_T$): We introduce a novel formulation, $\frac{\tanh\left(k(T-x_0)\right)}{1+c(T-x_{decay})^4}$, which captures the non-monotonic response of bacterial growth to temperature changes. This function combines a hyperbolic tangent term, representing the initial growth acceleration with temperature, and a quartic decay term, modeling the rapid decline in growth rate at high temperatures. This formulation presents a new and challenging form for LLM-based equation discovery methods, as it introduces operators and structures not commonly seen in the literature of this scientific context.
\item \textbf{pH Dependency} ($f_{\text{pH}}$): Our custom pH function, $\exp\left(-\left|\text{pH}-\text{pH}_{opt}\right|\right) \sin\left(\frac{(\text{pH}-\text{pH}_{min})\pi}{\text{pH}_{max}-\text{pH}_{min}}\right)^2$, combines exponential and trigonometric terms to model the complex relationship between bacterial growth and pH levels. This formulation captures both the optimal pH range for growth and the symmetric decline in growth rate as pH deviates from the optimum. It also poses challenging setting for LLM-based equation discovery methods with structures uncommon in the relevant scientific context.
\end{itemize}

\begin{figure}[t]
\centering
\includegraphics[width=0.8\linewidth]{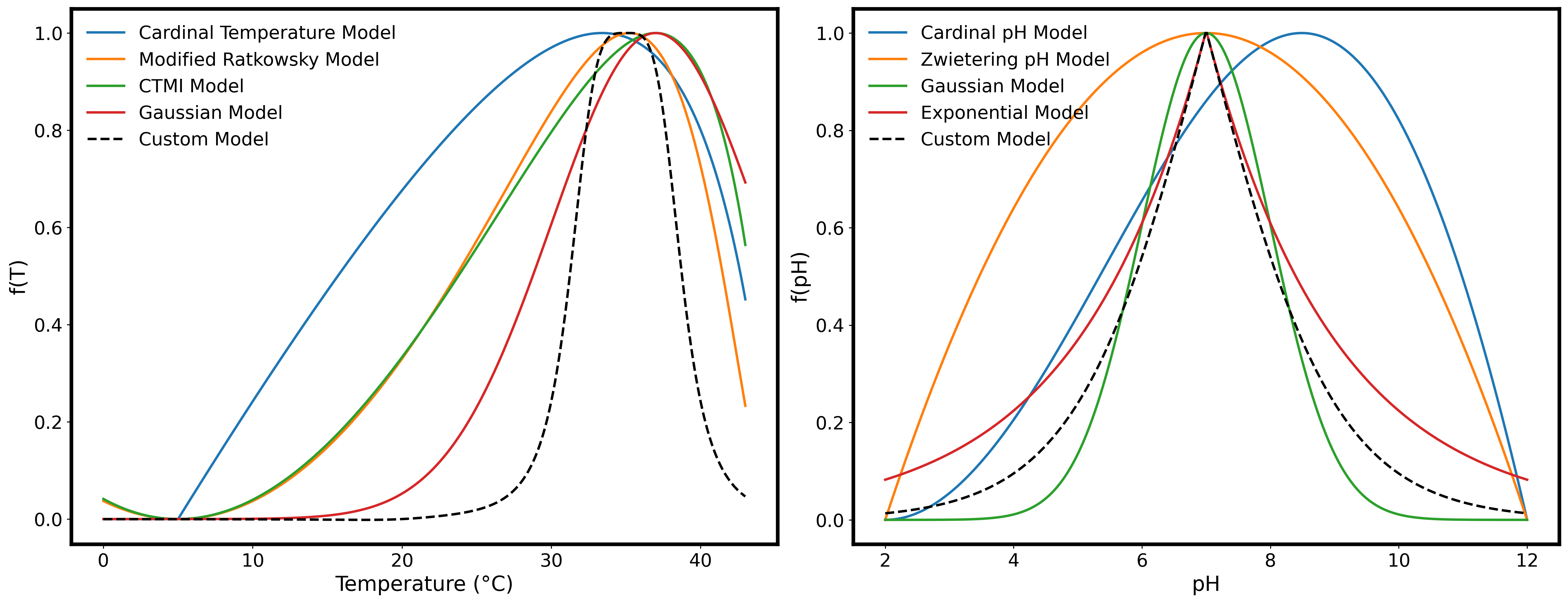}
\caption{\small Scheme of some established models from literature for temperature and pH impact in bacterial growth compared to our custom-designed model behavior.
}
\vspace{-0.5em}
\label{fig:pH_dependency}
\end{figure}

Fig.~\ref{fig:pH_dependency} illustrates the behavior of our custom-designed $f_T(T)$ and $f_{\text{pH}}(\text{pH})$ functions in comparison to established models from the literature for temperature and pH impact in bacterial growth. As observed, our custom models maintain key characteristics of bacterial growth responses while introducing complexities that challenge equation discovery methods. The temperature dependency model shows a sharper optimal peak and more rapid decline at high temperatures compared to traditional models, while the pH dependency model exhibits a narrower optimal range with steeper declines outside this range.

This carefully constructed benchmark serves multiple purposes: ($i$)~It leverages LLMs' prior knowledge of bacterial growth patterns and common mathematical functions used in biological modeling; ($ii$)~It prevents trivial LLM recitation by introducing novel combinations of functions and operators that go beyond standard models; and 
($iii$)~It challenges equation discovery systems to identify complex, biologically plausible relationships from data, simulating the process of scientific discovery.

\subsection{Material Stress Behavior Analysis}
\label{sec:app-stress}
\begin{figure}[t]
\centering\includegraphics[width=0.6\linewidth]{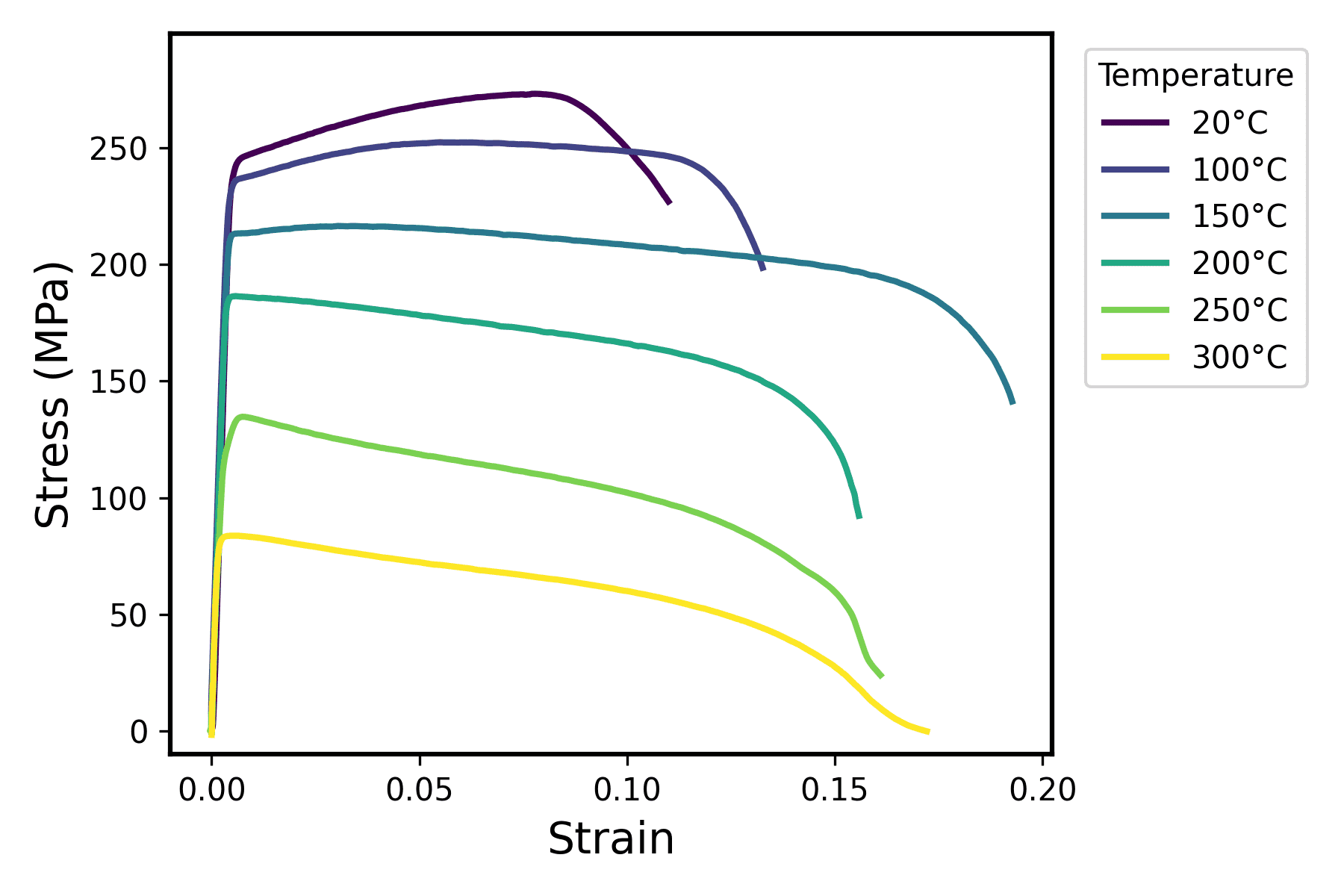}
\caption{\small Stress-strain curves of Aluminium 6061-T651 under various temperatures (data from \citep{stress-data-brief})}
\label{fig:stress_gt}
\vspace{-1em}
\end{figure}

The analysis of material stress behavior with regard to temperature variations is a critical area of study in materials science and engineering. Our focus on experimental data of Aluminium 6061-T651 \citep{stress-data-brief} provides a new experimental case study for LLM-based equation discovery methods. Fig.~\ref{fig:stress_gt} presents stress-strain curves for this alloy across a range of temperatures, offering rich insights into its mechanical behavior. For this problem, the experimental data represents the tensile behavior of material under uniaxial tension for 6 different temperatures from 20°C to 300°C. 
We allocate the data corresponding to $T=200^\circ$C for use as the out-of-domain validation set.

The stress-strain curves in Fig.~\ref{fig:stress_gt} reveal several key features:
\textbf{Temperature Dependence}: As temperature increases from 20°C to 300°C, we observe a significant decrease in both yield strength and ultimate tensile strength. 
\textbf{Elastic Region}: The initial linear portion of each curve represents the elastic region, where deformation is reversible. The slope of this region, known as Young's modulus, appears to decrease with increasing temperature, indicating reduced stiffness at higher temperatures. \textbf{Plastic Region}: Beyond the yield point, the curves exhibit non-linear behavior characteristic of plastic deformation. The shape of this region varies with temperature, suggesting changes in work hardening behavior. \textbf{Failure Region}: The endpoints of the curves indicate material failure. Notably, the strain at failure generally increases with temperature, implying enhanced ductility at higher temperatures. \textbf{Complex Non-linearity}: The overall shape of the curves exhibit piece-wise form, particularly in the plastic region, displays complex non-linear behavior that varies significantly with temperature and cannot be simply modeled with closed-form mathematical expressions.

The complex, nonlinear behavior of materials under varying experimental conditions poses significant challenges in developing comprehensive theoretical models.
Moreover, unlike many physics problems where equation forms might be known or suspected, stress-strain-temperature relationships for specific materials often lack a universally accepted theoretical model. This absence necessitates a more exploratory, data-driven approach to empirical equation discovery.
By incorporating this real-world experimental materials science problem into our benchmark problems, we aim to evaluate the ability of \modelname to discover physically meaningful and interpretable equations in a domain where empirical modeling often dominates 

\begin{figure}[t]
\centering
\includegraphics[width=0.9\linewidth]{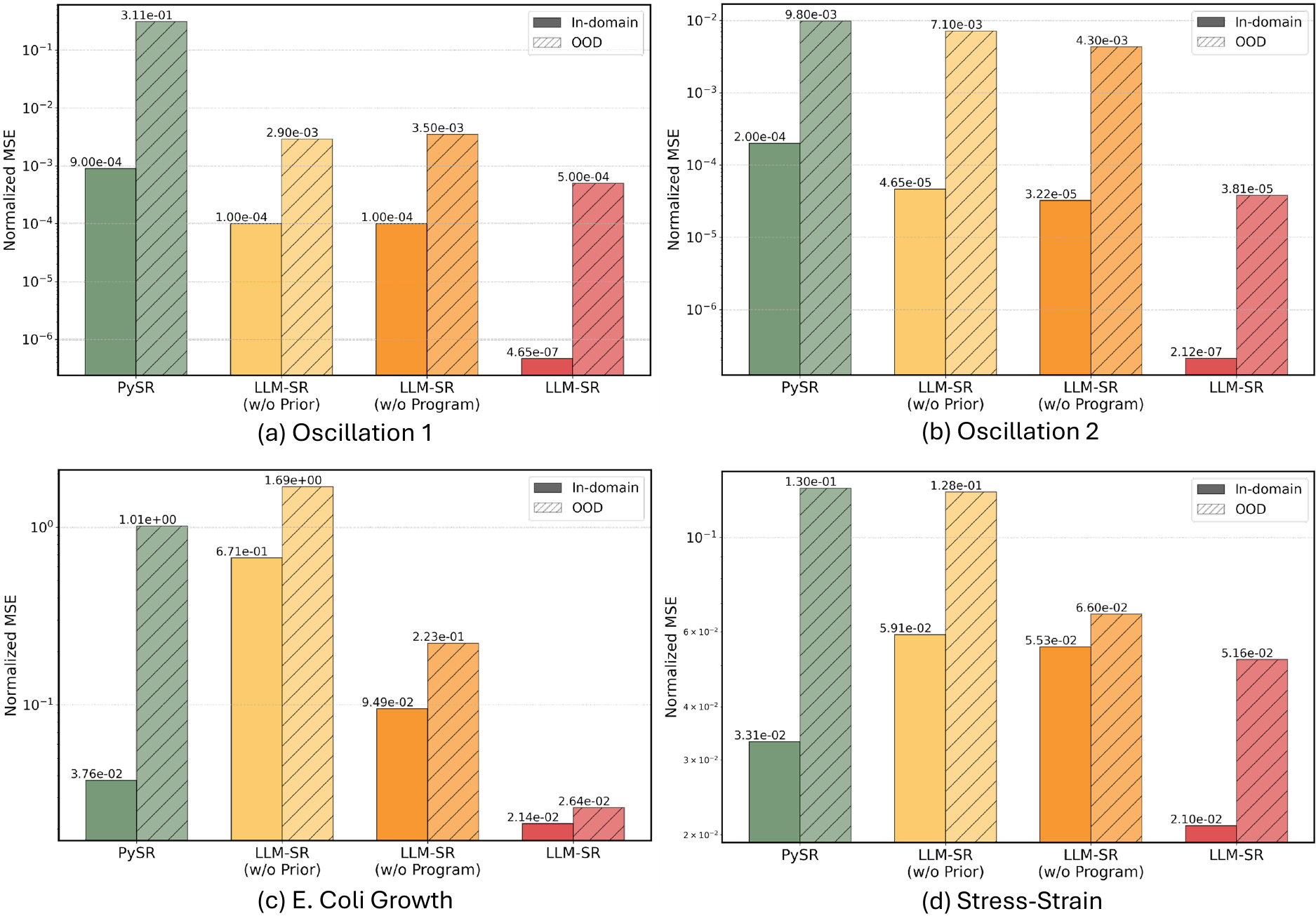}
\caption{\small Ablation comparison of \modelname variants (\texttt{GPT-3.5} backbone) and PySR on four benchmarks, showing in-domain and out-of-domain errors for \textbf{(a)}~Oscillation 1, \textbf{(b)}~Oscillation 2, \textbf{(c)}~E.coli Growth, and \textbf{(d)}~Stress-Strain problems. \modelname consistently outperforms PySR and ablated versions, highlighting the importance of both prior knowledge and equation program representation components.
}
\vspace{-0.5em}
\label{fig:ablation_pp}
\vspace{-0.5em}
\end{figure}

\vspace{-0.8em}
\section{Additional Ablation Study}
\label{sec:app-ablation}
\vspace{-0.8em}
To evaluate the key components of our \modelname framework, we conducted additional ablation studies across all benchmark datasets. These experiments specifically target two main motivations behind \modelname: (1)~leveraging scientific prior knowledge and (2)~utilizing code generation capabilities of LLMs in the context of scientific equation discovery.
We present the results of these ablation studies in Fig.~\ref{fig:ablation_pp}, which compares the performance of three \modelname variant models: ($i$)~\modelname~(\textit{w/o Prior}): This variant removes the incorporation of problem-specific prior knowledge from the LLMs' input prompt; ($ii$)~\modelname~(\textit{w/o Program}): This variant eliminates the use of program representation for hypotheses; and ($iii$)~\modelname: This variant is the final version of \modelname model including all components. 
We evaluated these models on all benchmark problems. For each problem, we assessed performance under both in-domain and out-of-domain (OOD) test settings, providing a robust evaluation of generalization capabilities in discovered equations. 

Fig.~\ref{fig:ablation_pp} shows that both ablated components are critical to \modelname's success. Across all datasets and in both test settings, \textit{w/o Prior} and \textit{w/o Program} consistently underperformed compared to the full \modelname model.
In both Oscillation 1 and Oscillation 2 problems, prior knowledge and code generation capabilities demonstrated comparable impacts on model performance. This held true for both in-domain and OOD settings, suggesting that both components contribute similarly to the model's understanding and generalization of oscillatory systems.
For the E. coli Growth problem, we observed a more pronounced effect of prior knowledge. 
This indicates that domain-specific knowledge plays a particularly crucial role in modeling for this problem.
Interestingly, while both components had similar impacts on in-domain performance in stress-strain problem, prior knowledge appeared to have a more substantial effect on OOD performance. This aligns with our intuition that OOD generalization should correlate more strongly with prior domain knowledge of the system.

\begin{table}[h]
\centering
\resizebox{0.6\linewidth}{!}{
\begin{tabular}{lcccc}
\toprule
\multirow{2}{*}{Model} & \multicolumn{2}{c}{Stress-Strain} & \multicolumn{2}{c}{Oscillation 2} \\
\cmidrule(lr){2-3} \cmidrule(lr){4-5}
& ID$\downarrow$ & OOD$\downarrow$ & ID$\downarrow$ & OOD$\downarrow$ \\
\midrule
\modelname~(w/o multi-island \& sampling) & 0.0257 & 0.1010 & 6.23e-6 & 0.0008 \\
\modelname & \textbf{0.0210} & \textbf{0.0516} & \textbf{2.12e-7} & \textbf{3.81e-5} \\
\bottomrule
\end{tabular}
}
\caption{\small Impact of multi-island design on \modelname performance (with GPT-3.5 backbone) across two benchmark problems measured by Normalized Mean Squared Error.}
\label{table-island}
\end{table}

We further conducted ablation experiments to analyze the impact of the multi-island buffer design and corresponding sampling strategy. Table~\ref{table-island} compares the performance of \modelname with and without the multi-island components on two benchmark problems. Specifically, we evaluated a variant that uses only one island in the buffer and employs a simple deterministic top-k selection approach for in-context example selection. The results demonstrate that the multi-island design positively impacts \modelname performance in both in-domain and OOD settings. Qualitatively, we observe that the number of islands plays a critical role in balancing exploitation and exploration. With fewer islands, the framework exhibits reduced exploration capabilities, generating less diverse equation hypotheses and converging prematurely to equation structures produced in early iterations.

\section{Additional Experiments}
\subsection{Noise Robustness Analysis}
\label{sec:app-noise}

To evaluate the robustness of \modelname to imperfect data conditions commonly encountered in real-world settings, we conducted a systematic analysis of model performance under varying levels of noise. We specifically focused on the Oscillation 2 benchmark problem, introducing controlled Gaussian noise with different standard deviations ($\sigma = \{0, 0.01, 0.05, 0.1\}$) to the training data. This analysis provides insights into how the incorporation of prior knowledge through \modelname affects equation discovery performance when dealing with noisy measurements.

\begin{figure}[ht]
\centering
\includegraphics[width=0.7\linewidth]{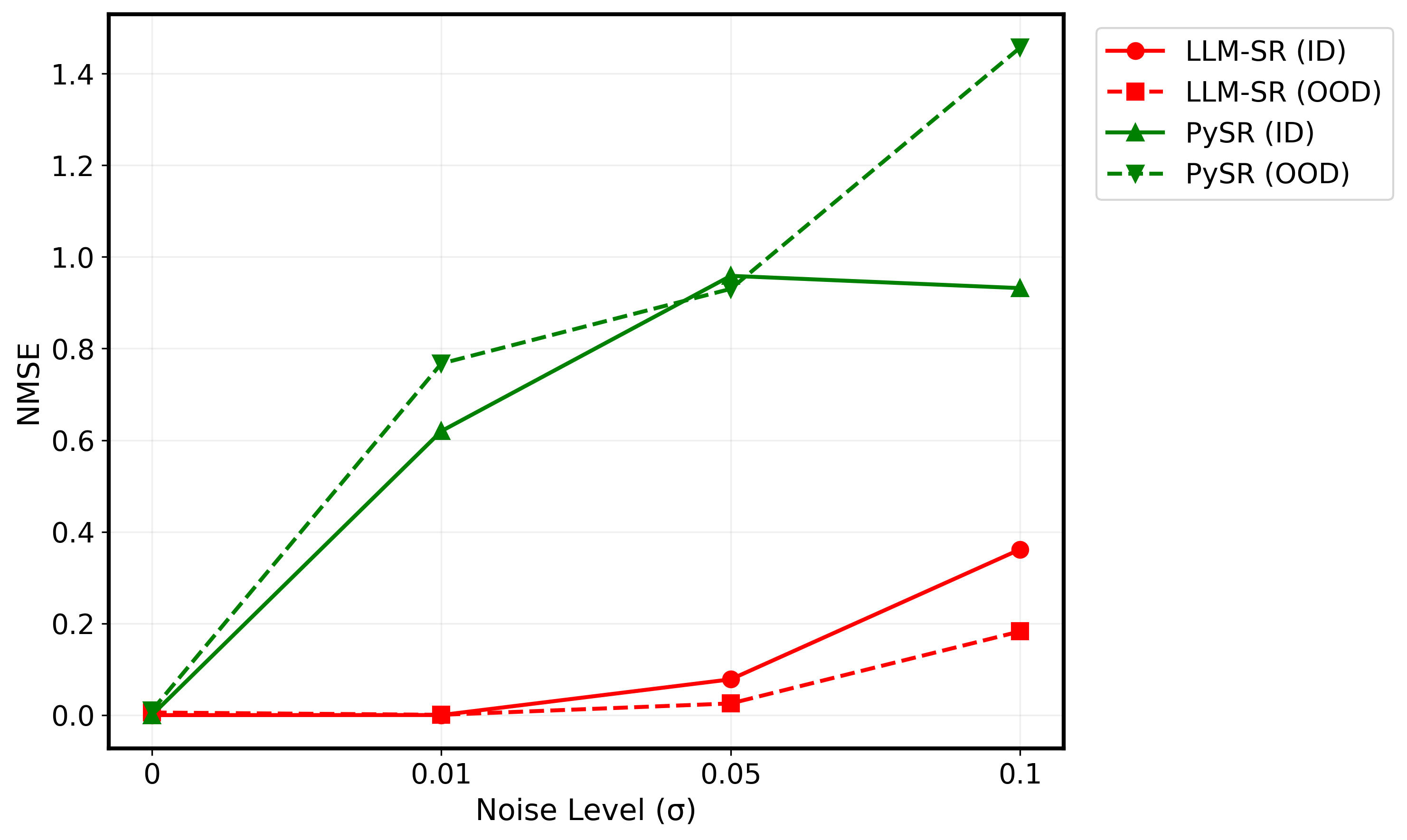}
\caption{Noise robustness analysis of \modelname compared to PySR on the Oscillation 2 benchmark. The plot shows normalized mean squared error (NMSE) for both in-domain and out-of-domain (OOD) predictions under different levels of Gaussian noise ($\sigma = {0, 0.01, 0.05, 0.1}$). While performance degrades with increasing noise for both methods, \modelname maintains better robustness, particularly for OOD predictions, demonstrating the value of incorporating domain knowledge in noisy real-world settings.
}
\vspace{-0.5em}
\label{fig:noise}
\vspace{-0.5em}
\end{figure}

Fig.~\ref{fig:noise} presents a comparative analysis between \modelname and PySR under different noise conditions. The results demonstrate that while increasing noise levels generally degrade performance across all methods, \modelname exhibits notably better resilience to noise compared to traditional approaches. Specifically, at moderate noise levels ($\sigma = 0.01$, $\sigma = 0.05$), \modelname maintains significantly lower NMSE compared to PySR, particularly in out-of-domain predictions.

This enhanced robustness can be attributed to the incorporation of domain knowledge through LLM prompting, which helps constrain the search space to physically plausible solutions even in the presence of noise. These findings highlight an important aspect of symbolic regression: as data quality decreases, the value of incorporating prior knowledge becomes increasingly significant. This observation aligns with the broader principle that when evidence (data) becomes less reliable, the role of priors in inference becomes more crucial.

\subsection{LLM-based optimization baselines}
\label{sec:app-llmopt}

Several frameworks have emerged recently exploring the integration of LLMs into optimization tasks \citep{LMX-Lehman-2023, LLM-Optimizer, FunSearch}. Among these works, LMX~\citep{LMX-Lehman-2023} has included symbolic regression as one of its experimental tasks, though primarily as a proof-of-concept rather than aiming to achieve state-of-the-art performance. Their implementation does not incorporate domain-specific prior knowledge, generates complete equations with LLM instead of optimizable skeletons, uses older LLM models (Galactica, Pythia), and lacks several design elements present in \modelname such as equation-as-program representation and multi-island dynamic memory management for diverse exploration. We conducted experiments to evaluate \modelname alongside LMX and FunSearch~\citep{FunSearch}. For LMX, we utilized their open-source implementation directly, while for FunSearch, we adapted their prompt and feedback design to suit the equation-as-program task. All methods were run for the same number of iterations as \modelname, using GPT-3.5 as the LLM backbone. 

\begin{table}[h]
\centering
\resizebox{0.60\columnwidth}{!}{
\begin{tabular}{lcccc}
\toprule
\multirow{2}{*}{Model} & \multicolumn{2}{c}{Oscillation 1} & \multicolumn{2}{c}{Oscillation 2} \\
\cmidrule(lr){2-3} \cmidrule(lr){4-5}
& ID$\downarrow$ & OOD$\downarrow$ & ID$\downarrow$ & OOD$\downarrow$ \\
\midrule
DSR & 0.0087 & 0.2454 & 0.0580 & 0.1945 \\
uDSR & 0.0003 & 0.0007 & 0.0032 & 0.0015 \\
PySR & 0.0009 & 0.3106 & 0.0002 & 0.0098 \\
\hline
LMX & 0.5031 & 48.93 & 1.004 & 0.9371 \\
FunSearch & 0.4840 & 8.059 & 0.7234 & 0.5861 \\
\hline
\modelname~(w/o skeleton+optimizer) & 0.1371 & 0.6764 & 0.3780 & 0.3751 \\
\modelname~(w/o Prior) & 0.0001 & 0.0029 & 4.65e-5 & 0.0071 \\
\modelname~(w/o Program) & 0.0001 & 0.0035 & 3.22e-5 & 0.0043 \\
\modelname & \textbf{4.65e-7} & \textbf{0.0005} & \textbf{2.12e-7} & \textbf{3.81e-5} \\
\bottomrule
\end{tabular}}
\vspace{-0.5em}
\caption{\small Additional experimental results comparing different optimization methods on oscillator problems measured by Normalized Mean Squared Error.}
\vspace{-0.5em}
\label{table-llmopt}
\end{table}


Results in Table~\ref{table-llmopt} show that \modelname and its variants achieve better performance than LMX and FunSearch in both in-domain and OOD test settings. Even when the skeleton+optimizer design is ablated (allowing LLM to generate complete equations without placeholder parameters), performance drops significantly but still outperforms both LMX and FunSearch. Further ablation of prior knowledge or program representation components shows less dramatic impact, though the full \modelname incorporating all components achieves the best performance overall. Notably, traditional SR baselines (DSR, uDSR, PySR) also outperform the existing LLM-based approaches on both datasets, supporting our focus on comparing against state-of-the-art SR methods in the main experiments.

\subsection{LLM backbones}
\label{sec:app-llmback}

We have also conducted experiments with larger LLMs as the backbone for \modelname (than current Mixtral and GPT-3.5 backbones). Due to the computational budget limitations, we only conducted experiments on the oscillation 1 and oscillation 2 datasets. For a fair comparison, we ran this GPT-4o backbone model for the same number of iterations (2500) as other \modelname runs. Table~\ref{table-gpt4o} shows that on both datasets, LLM-SR (with GPT-4o) provides slightly better performance than LLM-SR with GPT-3.5 and Mixtral backbones, particularly in the OOD settings (reported numbers are NMSE as in Table~\ref{table-result}). These findings align with expectations that improved LLMs with better knowledge, reasoning, and programming capabilities have potential to enhance performance in LLM-SR framework.

\begin{table}[h]
\centering
\resizebox{0.50\columnwidth}{!}{
\begin{tabular}{lcccc}
\toprule
\multirow{2}{*}{Model} & \multicolumn{2}{c}{Oscillation 1} & \multicolumn{2}{c}{Oscillation 2} \\
\cmidrule(lr){2-3} \cmidrule(lr){4-5}
& ID$\downarrow$ & OOD$\downarrow$ & ID$\downarrow$ & OOD$\downarrow$ \\
\midrule
\modelname (\texttt{Mixtral}) &\cellcolor{mygray}\textbf{7.89e-8}& 0.0002 & 0.0030 & 0.0291 \\
\modelname (\texttt{GPT-3.5})& 4.65e-7 & 0.0005 & 2.12e-7 & 3.81e-5 \\
\modelname (\texttt{GPT-4o}) & 7.29e-6 & \cellcolor{mygray}\textbf{6.75e-5} & \cellcolor{mygray}\textbf{4.27e-10} & \cellcolor{mygray}\textbf{1.29e-6} \\
\bottomrule
\end{tabular}}
\vspace{-0.5em}
\caption{\small Additional experimental results comparing different LLM backbones on oscillator problems measured by Normalized Mean Squared Error.}
\vspace{-0.5em}
\label{table-gpt4o}
\end{table}

\section{Additional Qualitative Results}
\label{sec:app-results}

\paragraph{Discovery Trajectory}
In this section, we evaluate the progress of generated equations using \modelname over iterations. This analysis can illustrate the qualitative evolutionary refinement of the discovered equations.

\begin{figure}[t]
\centering
\includegraphics[width=0.65\linewidth]{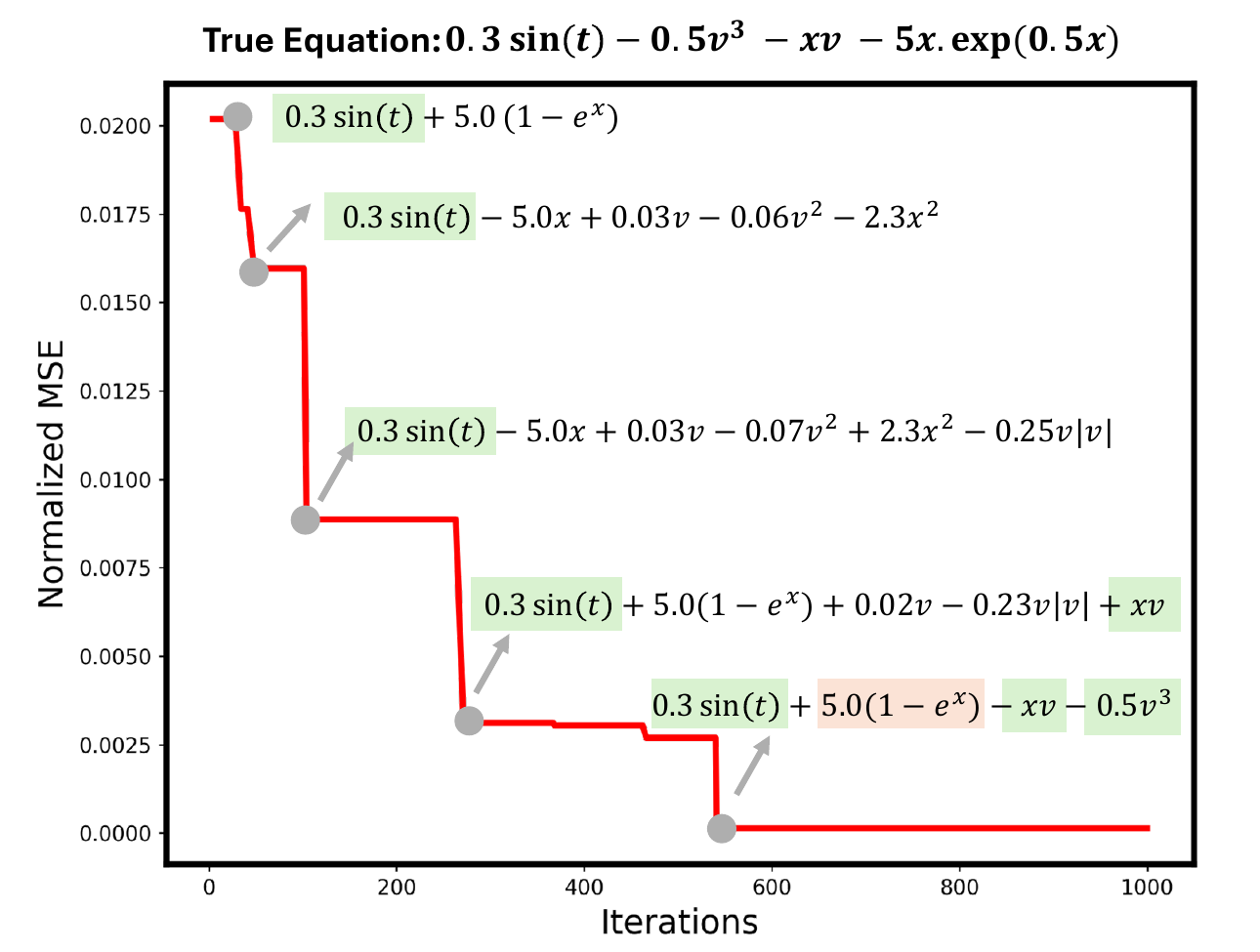}
\caption{\small 
Performance Trajectory of \modelname (\texttt{GPT-3.5}) along with the best-scoring simplified equations (after parameter optimization) over iterations on the Oscillation 2 problem. Green highlights indicate recovered symbolic terms from the true equation. 
}
\label{fig:osc2-annotated}
\end{figure}

Fig.~\ref{fig:osc2-annotated} shows the NMSE values for the Oscillation 2 dataset. For simplicity, we have provided the simplified equation versions of programs with their optimized parameters. We observe that some of the common nonlinear terms such as sinisoidal driving force term are found early in the search, while more complicated nonlinear terms are found later in the search. 
An interesting observation here is that while ground truth equation for this dataset is
$
\dot{v} = 0.3 \sin(t) -0.5 v^3 - x\cdot v -5x \cdot \exp (0.5 x),
$
\modelname has discovered the equation $
\dot{v} = 0.3 \sin(t) -0.5 v^3 - x\cdot v + 5 (1- \exp (x))
$ at the end.
By evaluating the different terms in these two forms, we observe that in fact $5(1 - \exp (x)) \approx -5x \cdot \exp (0.5 x)$ for $x \in (-2,2)$, which is the approximate range of displacement in this dataset.

\begin{figure}[!ht]
\centering
\includegraphics[width=\linewidth]{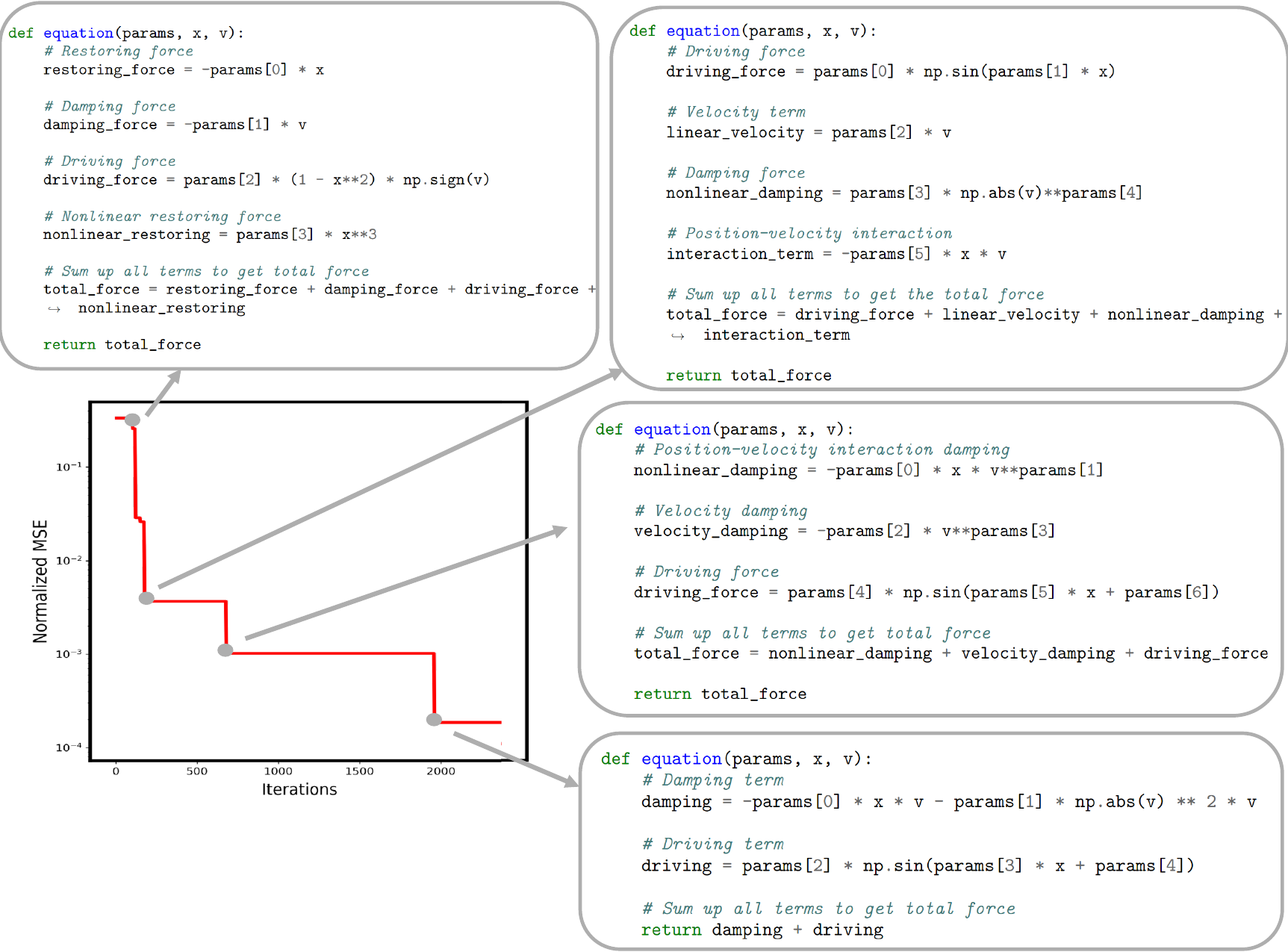}
\caption{\small 
Performance Trajectory of \modelname (\texttt{Mixtral}) along with the best-scoring equation program skeletons (before parameter optimization) over iterations on the Oscillation 1 problem.
}
\label{fig:osc1-annotated}
\end{figure}

Fig.~\ref{fig:osc1-annotated} illustrates the evolution of equation program skeletons for the Oscillation 1 problem. It can be observed that the model attempts to incorporate various nonlinear terms corresponding to driving, restoring, and damping forces, as evidenced by comments or variable names within the code, aiming to enhance accuracy. 

\begin{figure}[t]
\centering
\includegraphics[width=\linewidth]{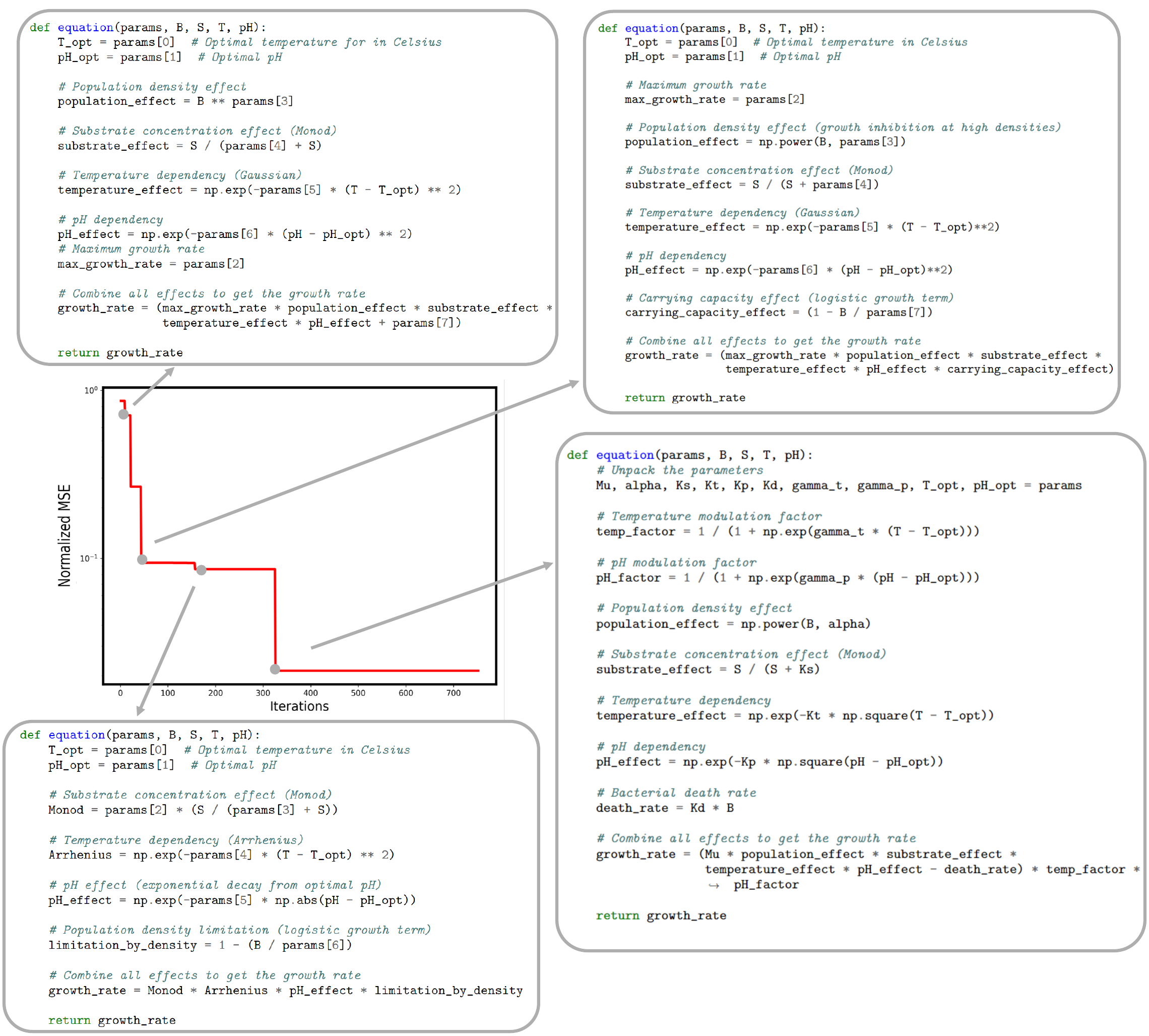}
\caption{\small 
Performance Trajectory of \modelname (\texttt{GPT-3.5}) along with the best-scoring equation program skeletons (before parameter optimization) over iterations on the E. coli growth problem.
}
\label{fig:bg-annotated}
\end{figure}

Similarly, Fig.~\ref{fig:bg-annotated} presents an annotated performance curve illustrating \modelname's performance on the E. coli growth rate equation discovery benchmark problem. It becomes apparent that the model recognizes the potential presence of optimal values for temperature and pH
from the early iterations which comes from model prior knowledge about the common bell-shaped effect of these variables on the growth rate (check Fig.~\ref{fig:pH_dependency}). To enhance accuracy, the model necessitates exploration and incorporation of various nonlinear forms. Notably, \modelname directs its evolutionary changes towards the more critical and variable aspects of the problem, specifically the pH and temperature effects, as opposed to other components such as substrate concentration represented by the Monod equation $\frac{S}{K+S}$. Additionally, the figure demonstrates \modelname's comprehension that different components of the function should be multiplied together in the final step, underscoring how prior domain knowledge of the problem structure can guide \modelname's evolutionary steps.

\begin{figure}[t]
\centering
\includegraphics[width=\linewidth]{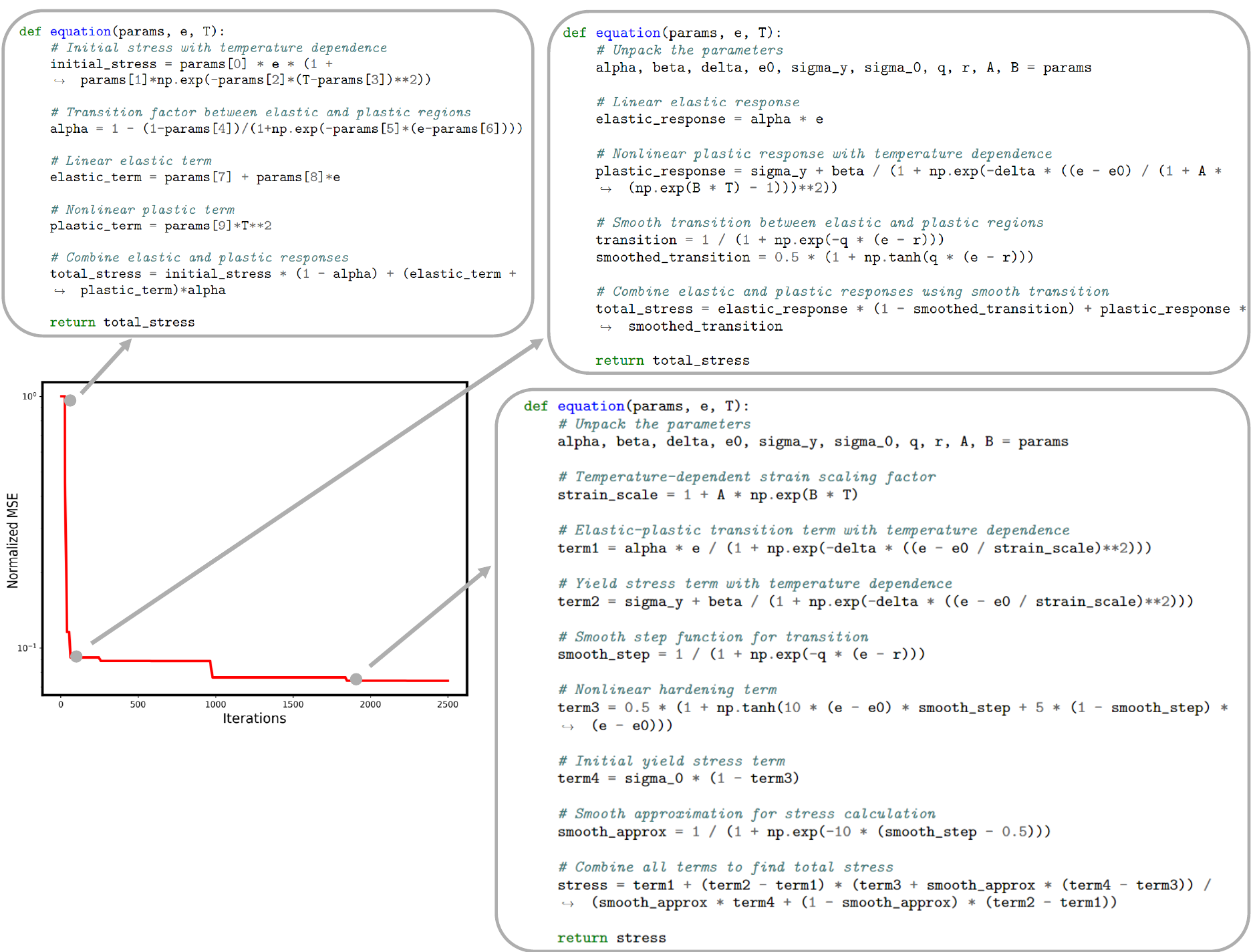}
\caption{\small 
Performance Trajectory of \modelname (\texttt{Mixtral}) along with the best-scoring equation program skeletons (before parameter optimization) over iterations on the Stress-Strain problem.
\vspace{-0.5em}
}
\vspace{-0.5em}
\label{fig:ss-annotated}
\end{figure}

Fig.~\ref{fig:ss-annotated} displays three distinct equation skeleton programs discovered by \modelname for the stress-strain problem over search iterations. As in previous cases, we notice the model's enhancement through exploration and incorporation of various nonlinear terms into the equations, particularly in the plastic region terms. An additional significant observation for this problem is that stress-strain relationships often exhibit piece-wise behavior (as it can also be observed in Fig.~\ref{fig:stress_gt}), which closed-form expression structures in traditional symbolic regression models struggle to model. However, \modelname represents equation skeletons as programs, thus, it can employ conditional rules (If-Else) or their continuous relaxations, utilizing step-wise nonlinear differentiable functions such as the \texttt{sigmoid} function to model smooth piece-wise relations. This differentiability and smooth approximation of if-else conditions are particularly helpful for the parameter optimization step, providing smooth functions for the optimizer to navigate.

\begin{figure}[t]
\centering
\includegraphics[width=0.7\linewidth]{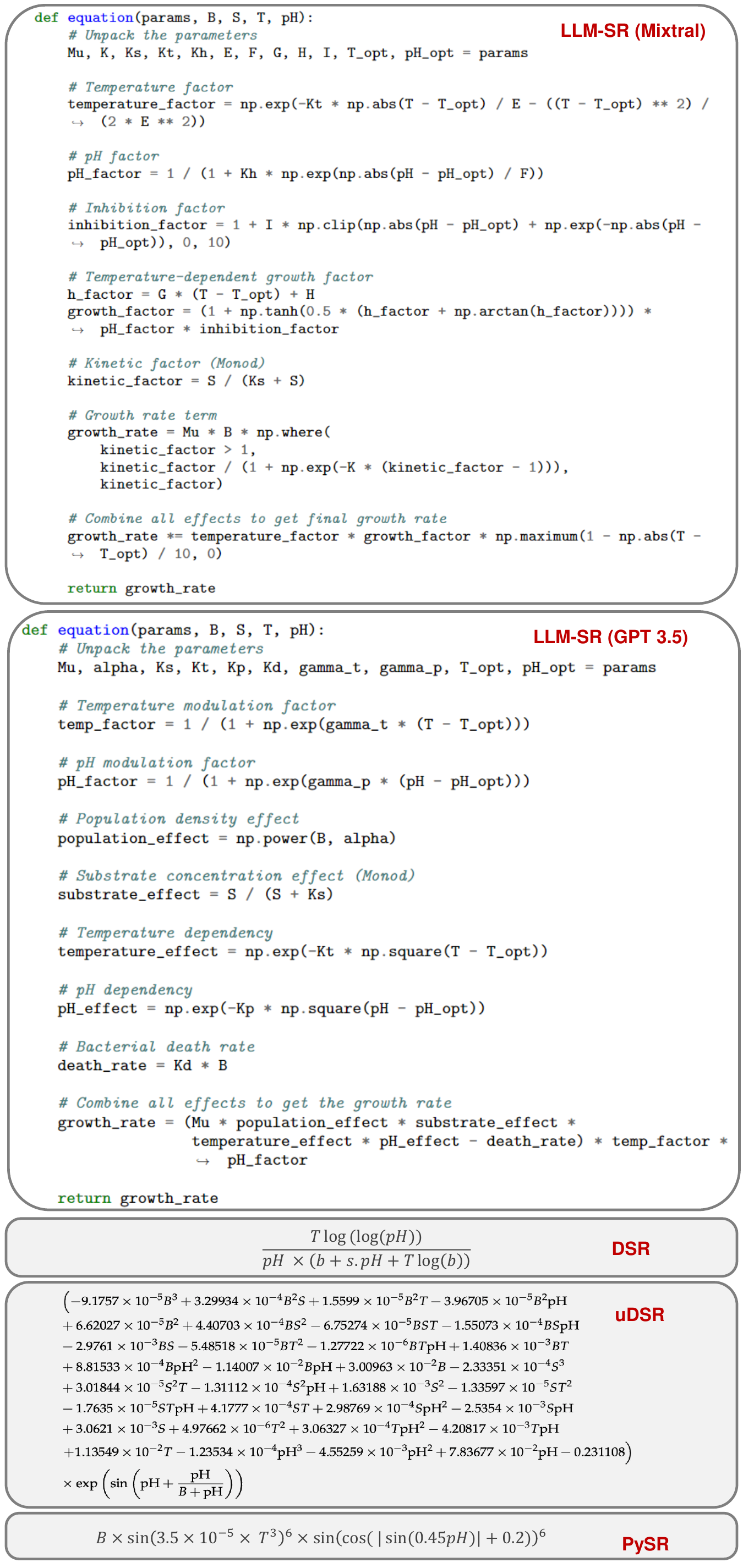}
\caption{\small 
Final discovered equations from \modelname and other leading SR baseline methods (DSR, uDSR, PySR for E. coli bacterial growth rate problem.
}
\vspace{-0.5em}
\label{fig:bacterial_equations}
\end{figure}

\begin{figure}[t]
\centering
\includegraphics[width=0.7\linewidth]{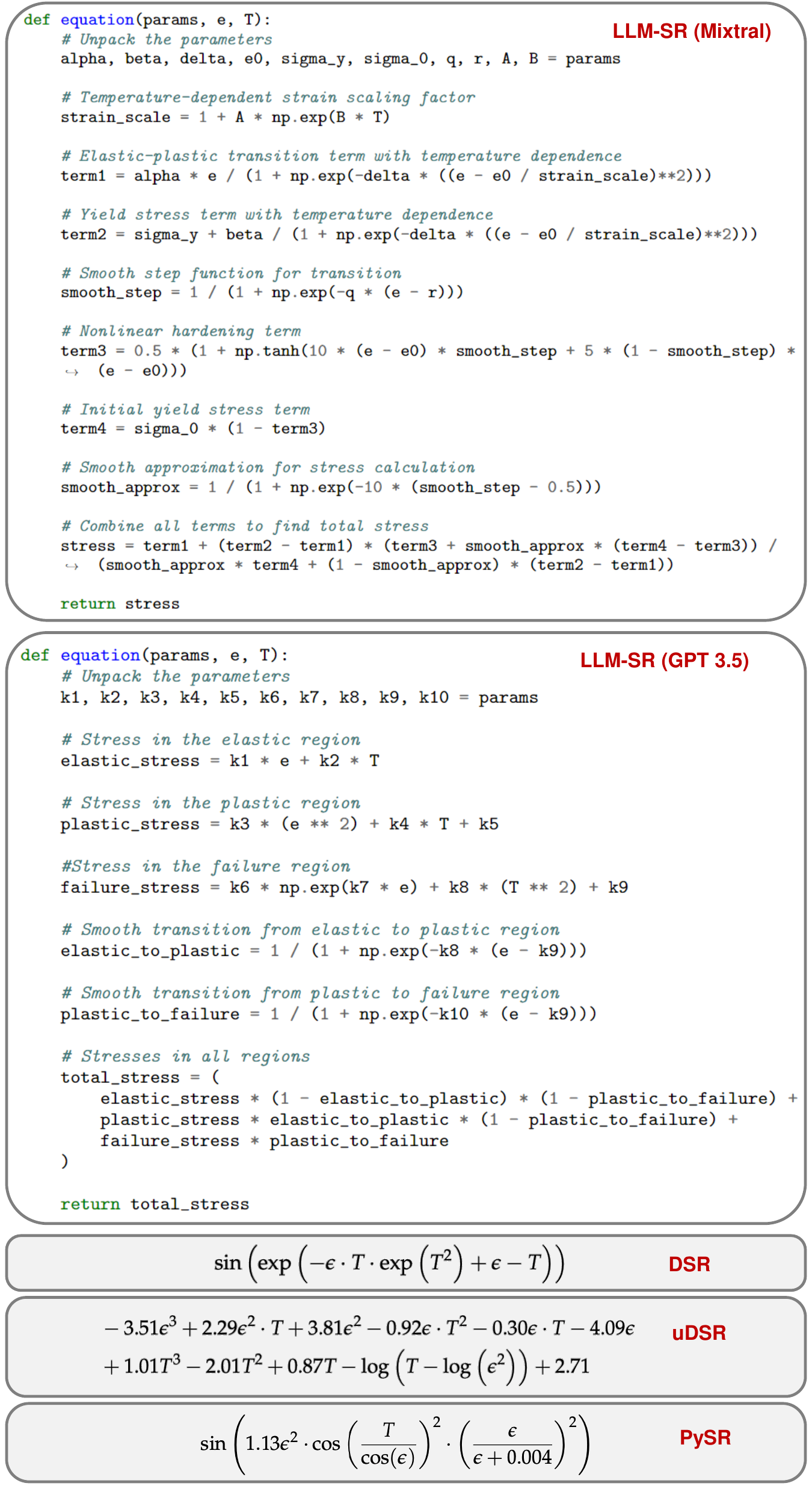}
\caption{\small 
Final discovered equations from \modelname and leading SR baseline methods (DSR, uDSR, PySR) for Stress-Strain problem.}
\vspace{-0.5em}
\label{fig:stress_equations}
\end{figure}

\paragraph{Discovered Equations}
Fig.~\ref{fig:bacterial_equations} and Fig.~\ref{fig:stress_equations} depict the equation programs identified by \modelname and other leading symbolic regression baselines (DSR, uDSR, and PySR) for the E. coli growth and the stress-strain problems, respectively. The diverse range of equation forms identified by different symbolic regression methods reflects the challenges posed by these datasets. Notably, in both datasets, the SR methods yield either lengthy or highly nonlinear equations that are not aligned with the prior knowledge of the systems, as evidenced by their poor out-of-domain (OOD) performance scores in Table \ref{table-result}. In contrast, \modelname finds flexible equation programs that are more interpretable and aligned with the domain-specific scientific priors of the systems.

\begin{figure}[t]
\centering
\includegraphics[width=0.8\linewidth]{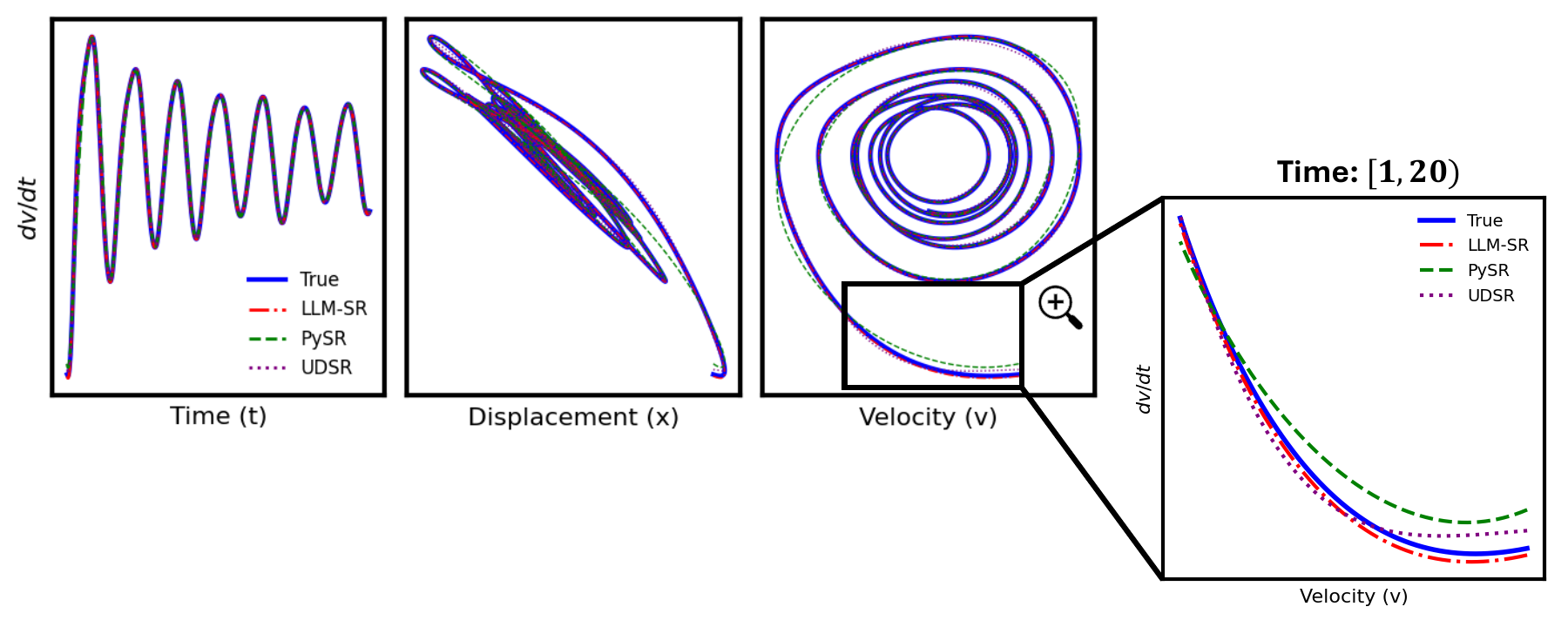}
\caption{\small Qualitative evaluation of the performance of \modelname on Oscillation 2 problem compared to uDSR and PySR baselines. Plots show the target acceleration with respect to time, displacement, and velocity. 
}
\vspace{-0.5em}
\label{fig:qual_osc2}
\end{figure}

\begin{figure}[t]
\centering
\includegraphics[width=0.8\linewidth]{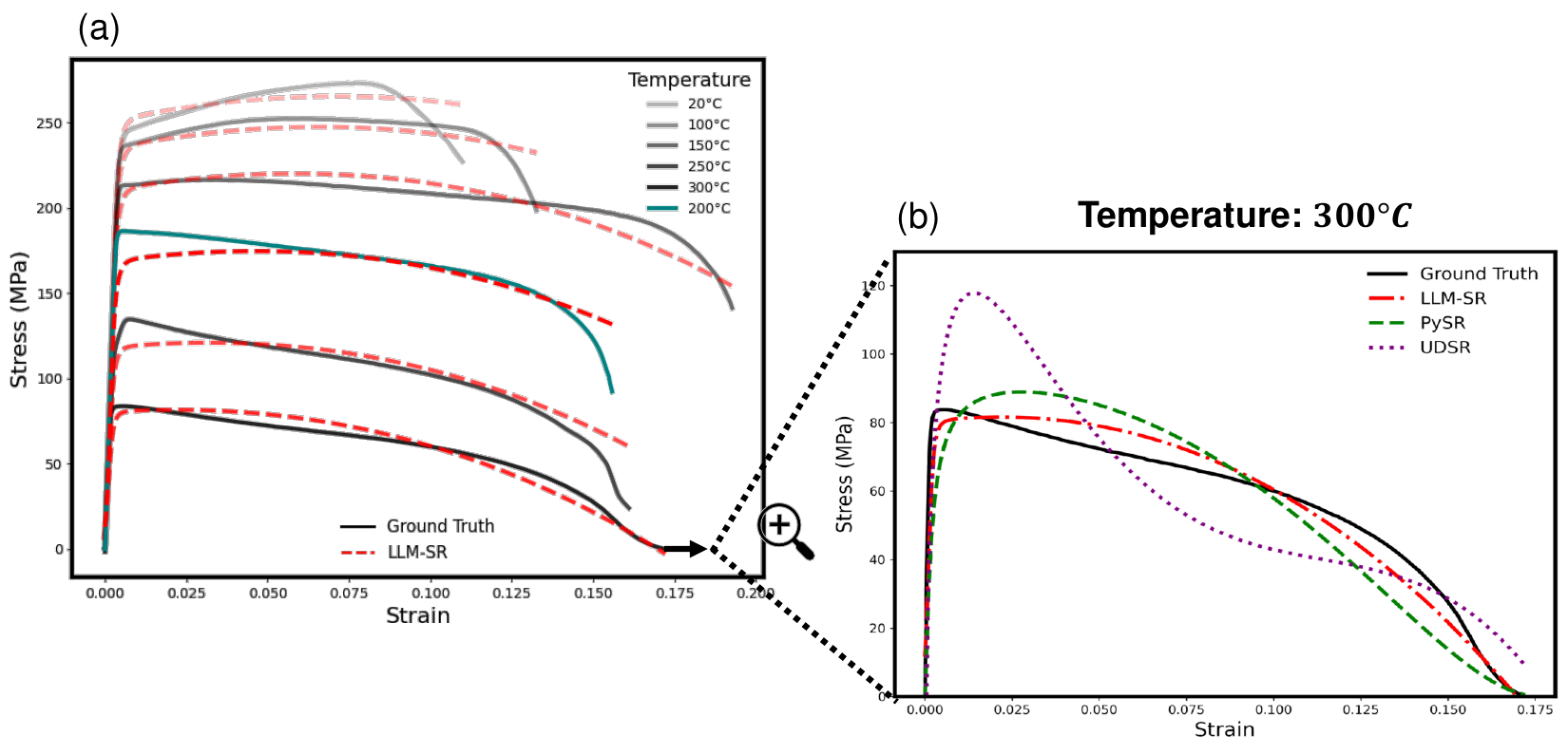}
\caption{\small Qualitative evaluation of 
\modelname performance for Stress-Strain problem compared to uDSR and PySR baselines.  Plots show the target stress with respect to strain and temperature. 
}
\vspace{-0.5em}
\label{fig:qual_stress}
\vspace{-0.5em}
\end{figure}

\paragraph{Behavior of Discovered Models}
Fig.~\ref{fig:qual_osc2} and Fig.~\ref{fig:qual_stress} offer a qualitative comparison by visually presenting the outputs of the equations obtained using \modelname, PySR, and uDSR. Upon examination, it becomes evident that the predictions generated by \modelname exhibit a notable degree of alignment with the ground truth data.
This alignment suggests that \modelname effectively captures the underlying patterns and dynamics of the data.

\end{document}